\documentclass{article}

\PassOptionsToPackage{compress}{natbib}  
\usepackage[final]{corl_2026}       

\usepackage{graphicx}
\usepackage{wrapfig}
\usepackage{placeins}
\usepackage{needspace}
\usepackage{caption}
\usepackage{tcolorbox}
\usepackage{amsmath}
\usepackage{amssymb}
\usepackage{booktabs}
\usepackage{multirow}
\usepackage{colortbl}

\usepackage{enumitem}
\usepackage{bm}
\usepackage[table]{xcolor}

\definecolor{lightcyan}{rgb}{0.88, 1.00, 1.00}

\newcommand{\acc}[2]{#1\,{\scriptsize$\pm$\,#2}}

\newcommand{\blfootnote}[1]{%
  \begingroup
  \renewcommand{\thefootnote}{}\footnote{#1}%
  \addtocounter{footnote}{-1}%
  \endgroup
}

\title{\texorpdfstring{%
  \makebox[0pt][r]{\raisebox{-0.45\height}{\includegraphics[height=2.1\baselineskip]{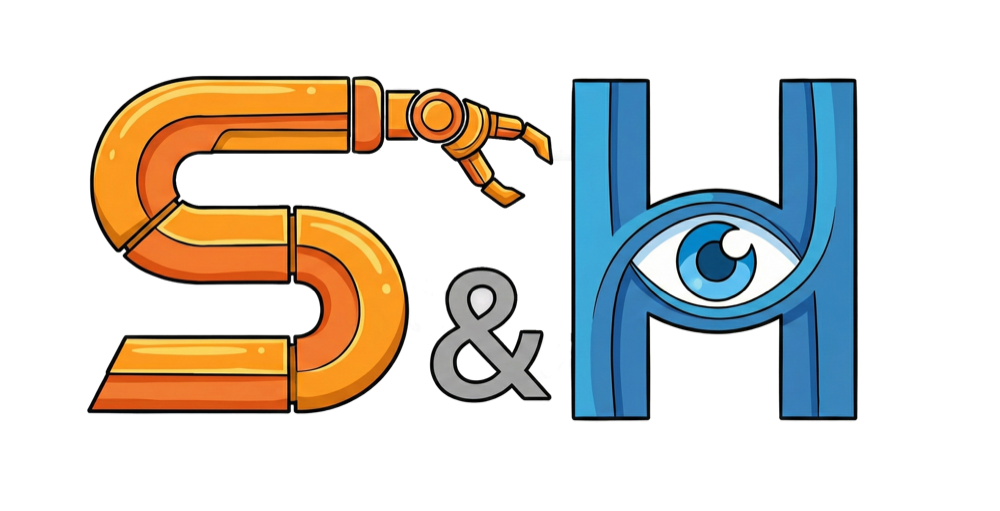}}\hspace{0.5em}}%
  \begin{tabular}[c]{@{}c@{}}Sim-and-Human Co-training for Data-Efficient\\ and Scene-Generalizable Bimanual Manipulation\end{tabular}%
}{Sim-and-Human Co-training for Data-Efficient and Scene-Generalizable Bimanual Manipulation}}

\author{
  Kaipeng Fang\textsuperscript{1},
  Weiqing Liang\textsuperscript{1},
  Yuyang Li\textsuperscript{1},
  Ji Zhang\textsuperscript{2},
  Pengpeng Zeng\textsuperscript{3}, \\
  \bfseries
  Heng Tao Shen\textsuperscript{3},
  Jingkuan Song\textsuperscript{3,4},
  Lianli Gao\textsuperscript{1\dag} \\[1.2ex]
  \normalfont
  \textsuperscript{1}University of Electronic Science and Technology of China \\
  \normalfont
  \textsuperscript{2}Southwest Jiaotong University \qquad
  \textsuperscript{3}Tongji University \qquad
  \textsuperscript{4}Shanghai Innovation Institute
}

\begin{document}
\maketitle
\blfootnote{\textsuperscript{\dag}: denotes corresponding author.}
\vspace{-2em}

\begin{figure}[!h]
  \centering
  \includegraphics[width=0.95\linewidth]{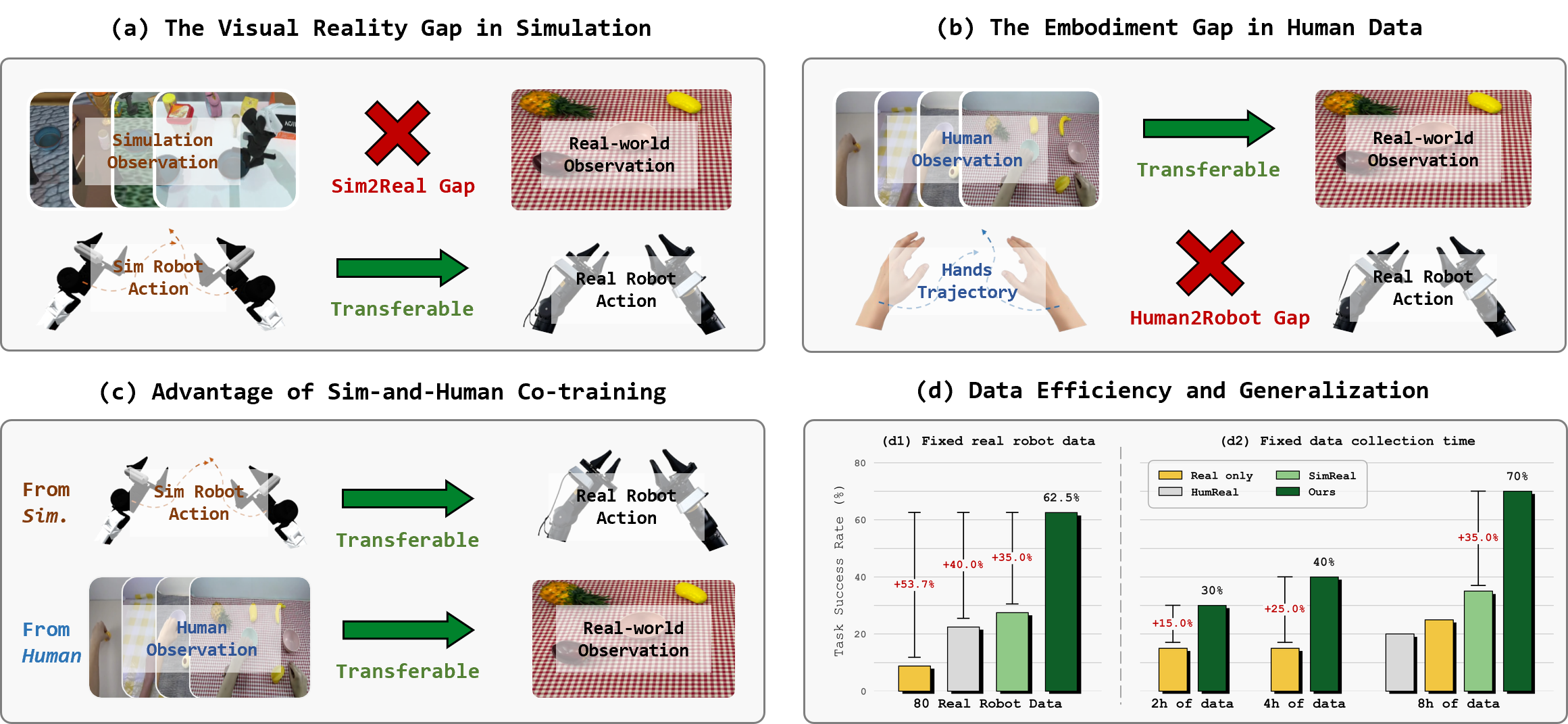}
  \caption{\textbf{Sim-and-Human Co-training.} \textbf{(a)} Simulation data offers transferable actions but suffers from the Sim2Real gap; \textbf{(b)} Human data provides realistic observations but is limited by the Human2Robot gap. \textbf{(c)} Our SimHum framework extracts kinematic priors from simulation and visual priors from human observations, with each prior compensating for the other source's gap. \textbf{(d)} SimHum achieves strong OOD generalization and data efficiency on bimanual manipulation.}
  \label{fig:teaser}
\end{figure}
\vspace{-1em}

\begin{abstract}
Real-robot demonstrations are prohibitively expensive, while simulation data and real-world human demonstrations are both scalable but each leaves a distinct gap—simulation suffers from a \textit{sim-to-real visual gap}, and human data suffers from a \textit{human-to-robot embodiment gap}. In this work, we identify a natural yet underexplored complementarity between these sources: simulation contributes robot-valid actions absent in human data, while human data provides real-world observations that simulation struggles to render. Building on this insight, we present \textbf{SimHum}, a co-training recipe that extracts kinematic priors from simulation and visual priors from human observations, then fine-tunes on a small real-robot dataset. SimHum exhibits strong scene-generalizable and data-efficient capabilities. With only $\mathbf{80}$ real-robot episodes per task, it achieves $\mathbf{62.5\%}$ success on held-out OOD scenes across four bimanual tabletop tasks, $\mathbf{53.7\%}$ higher than Real only in absolute success rate. Moreover, in a controlled data-collection study with matched collection time, SimHum improves over the best single-source pre-training baseline by $\mathbf{35.0\%}$ in absolute success rate. More information at our project page\footnote{Project page: \url{https://kaipengfang.github.io/sim-and-human/}}.
\end{abstract}


\section{Introduction}
\label{sec:intro}

A long-standing challenge in robot learning is enabling policies to generalize across complex manipulation tasks in the unconstrained real world. Recent advances in robotic foundation models~\citep{zhao2023act,chi2023diffusionpolicy, brohan2023rt1, brohan2023rt2,octo2024octo, kim2024openvla, liu2024rdt, black2024pi_0, intelligence2025pi_} suggest that large-scale pre-training is a promising path forward, but diverse robot data in the physical world is labor-intensive and prohibitively expensive to collect~\citep{padalkar2023openx, khazatsky2024droid, bu2025agibot, wu2025robocoin}.

Recent work has explored simulation data and human demonstrations as two scalable data sources that reduce reliance on real-robot data collection. Simulation can generate robot data with minimal human effort~\citep{raistrick2023infinigen, xu2023unidexgrasp, wang2023dexgraspnet, mandlekar2023mimicgen, mu2025robotwin1, chen2025robotwin, jiang2025dexmimicgen}, but learned policies must still overcome the sim-to-real gap~\citep{kumar2021rma, ha2020learning, Shirai2025dec, yu2022visual, qin2022dexpoint, cheng2025generalizable}; human demonstrations can be collected with ubiquitous consumer devices~\citep{qiu2025humanoid,tao2025dexwild, kareer2025egomimic}, but transferring them to robots remains challenging due to the human-to-robot kinematic gap. Prior works typically address these gaps through explicit alignment~\citep{qiu2025humanoid,tao2025dexwild,kareer2025egomimic,punamiya2025egobridge, liu2025immimic, yang2025egovla,lepert2025phantomtrainingrobotsrobots,lepert2025masqueradelearninginthewildhuman} or unified co-training~\citep{qiu2025humanoid,tao2025dexwild, kareer2025egomimic, wang2024hpt, nvidia2025gr00t, maddukuri2025simandreal, nasiriany2024robocasa,cheng2025generalizable}. Since both sources can be collected at scale without operating a real robot and are highly complementary---simulation provides robot-valid actions absent in human data (Figure~\ref{fig:teaser}a), while human data offers real-world observations that are difficult to synthesize in simulation (Figure~\ref{fig:teaser}b)---we pursue a different practical angle and ask: \textit{Can we co-train on simulation and human data to reduce the amount of real-robot data needed for bimanual manipulation?}

In this paper, we introduce a \textbf{Sim}-and-\textbf{Hum}an Co-training (\textbf{SimHum}) approach that jointly leverages simulation data (for robot-valid actions) and human demonstrations (for real-world observations). As shown in Figure~\ref{fig:teaser}(c), SimHum combines kinematic priors learned from simulation with visual priors learned from human data through joint pre-training, then fine-tunes on a small real-robot dataset, yielding a bimanual manipulation policy that generalizes to novel scenes. To this end, we first design a tailored data collection pipeline that establishes protocol-level kinematic alignment for simulation data and visual alignment for human data. We then jointly pre-train a modular diffusion policy on the two data sources, and restructure the pre-trained policy by selectively retaining the modules useful for real-robot fine-tuning while discarding components redundant for deployment.

We empirically evaluate SimHum on real-world bimanual manipulation (Section~\ref{subsec:evaluation tasks}). With only $\mathbf{80}$ real-robot episodes per task, SimHum achieves $\mathbf{62.5\%}$ success on held-out OOD scenes across four bimanual tabletop tasks, compared with $\mathbf{8.8\%}$ when training on the same real data alone (Figure~\ref{fig:teaser}d1). In a controlled data-collection study with matched collection time, reallocating half of the real-robot time to human data collection, with simulation generated in parallel, improves over the best single-source baseline by $\mathbf{35.0\%}$ in absolute success rate (Figure~\ref{fig:teaser}d2), suggesting that SimHum can further reduce the amount of real-robot data needed for bimanual manipulation beyond single-source pre-training. Our main contributions can be summarized as follows:
\begin{itemize}[itemsep=3pt,parsep=0pt,topsep=3pt,leftmargin=*]
    \item We present SimHum, a co-training framework that combines simulation data and human demonstrations to reduce the amount of real-robot data needed for bimanual manipulation.
    \item SimHum combines a data pipeline that aligns simulation kinematics and human visual observations with the target real-robot setup, and a modular diffusion policy that retains source-specific modules during real-robot fine-tuning.
    \item Across four real-world bimanual manipulation tasks, SimHum improves performance in both in-distribution and held-out OOD settings under limited real-robot data, and we further provide a decoupling analysis that characterizes the distinct contributions of simulation and human data to policy generalization.
\end{itemize}

\section{Related Works}
\label{sec:related}

\subsection{Imitation Learning for Robot Manipulation}
Imitation Learning (IL), particularly in the form of Behavior Cloning (BC), has emerged as a dominant paradigm for acquiring robust visuomotor policies directly from expert demonstrations~\citep{pomerleau1988alvinn, osa2018survey, argall2009survey, florence2022implicit, jang2022bcz, li2024behavior, shridhar2023peract, jiang2023vima, goyal2023rvt, belkhale2023hydra, wu2025policy}. By mapping sensory observations to control actions, IL circumvents the complexities of manual controller design and reward engineering required by reinforcement learning. Recent advancements have demonstrated the remarkable scalability of this paradigm, driven largely by the evolution of policy architectures~\citep{zhao2023act,chi2023diffusionpolicy,brohan2023rt1, brohan2023rt2,octo2024octo, kim2024openvla, liu2024rdt, black2024pi_0, intelligence2025pi_} and data collection systems~\citep{karamcheti2023noir,qin2023anyteleop,fu2024mobile,wu2024gello,iyer2024open,yang2024ace,ahn2024autort, li2024fastumi, cheng2024opentelevision,wang2024dexcap, ye2025bunny, ze2025twist}. However, the prohibitive cost of acquiring large-scale real-world data remains a fundamental bottleneck that limits the generalization of imitation learning policies. Foundation-scale efforts such as OpenVLA~\citep{kim2024openvla} and $\pi_0$~\citep{black2024pi_0} address this bottleneck by scaling model capacity and pre-training on multi-embodiment real-robot data. Our work takes an orthogonal direction: rather than scaling real-robot pre-training, we study how two complementary scalable alternatives---simulation and human demonstrations---can be jointly exploited before fine-tuning on limited real-robot data.

\subsection{Learning from Scalable Heterogeneous Sources}
Leveraging scalable data sources, particularly simulation~\citep{raistrick2023infinigen, xu2023unidexgrasp, wang2023dexgraspnet, mandlekar2023mimicgen, mu2025robotwin1, chen2025robotwin, jiang2025dexmimicgen, xiang2020sapien, makoviychuk2021isaac, freeman2021brax, gu2023maniskill2, tao2025maniskill, zakka2025mujoco} and human demonstrations~\citep{taheri2020grab, chen2021understanding, sener2022assembly101, grauman2022ego4d, liu2022hoi4d, khirodkar2023egohumans, fan2023arctic, banerjee2025hot3d, li2025openhumanvid, chen2025vidbot, fan2025motionmillion, hoque2025egodex}, has become a standard practice to mitigate the scarcity of real-robot data. To harness these heterogeneous sources, prior works generally follow two primary paradigms: explicit domain alignment and unified policy co-training.

\noindent\textbf{Explicit Domain Alignment.} To align heterogeneous data for robot execution, many works focus on explicitly bridging distribution shifts. Methods in this category typically achieve alignment either through specific training objectives, such as Optimal Transport~\citep{cheng2025generalizable, punamiya2025egobridge} or MixUp interpolation~\citep{liu2025immimic}, or by preprocessing data via retargeting~\citep{qiu2025humanoid, yang2025egovla} and visual editing~\citep{kareer2025egomimic, lepert2025phantomtrainingrobotsrobots, lepert2025masqueradelearninginthewildhuman}. These methods are effective but often add extra assumptions, task-specific preprocessing, or auxiliary objectives.

\noindent\textbf{Co-training for Unified Representation.} In contrast to explicit alignment, a growing body of literature employs unified architectures to directly co-train on heterogeneous data, aiming to learn an embodiment-agnostic representation~\citep{cheng2025generalizable, qiu2025humanoid,tao2025dexwild, kareer2025egomimic, wang2024hpt, nvidia2025gr00t, maddukuri2025simandreal, nasiriany2024robocasa}. By circumventing complex alignment designs, these methods effectively scale up policy learning with massive mixed datasets. These methods typically pair one scalable source with real-robot data---co-training simulation with real data to bridge the sim-to-real gap, or human data with real data to bridge the embodiment gap. \textbf{SimHum} instead co-trains simulation and human data directly with each other, leveraging their intrinsic \textit{complementarity} before real-robot adaptation without explicit retargeting, visual editing, or auxiliary domain-alignment losses.

\section{Method}
\label{sec:method}

In this section, we describe \textbf{SimHum}, including its aligned data collection pipeline, joint pre-training on simulation and human data, and policy restructuring for real-robot fine-tuning.

\subsection{Problem Formulation}
\label{sec:formulation}

We formulate robotic manipulation as a conditional sequence generation problem. Let $\mathcal{D}_{real} = \{(o_t, s_t, \mathbf{a}_{t:t+H})\}$ be a real-world robot dataset, comprising visual observations $o_t \in \mathbb{R}^{H_{\text{img}} \times W_{\text{img}} \times 3}$, proprioceptive states $s_t \in \mathbb{R}^{d_s}$, and the ground-truth action sequence $\mathbf{a}_{t:t+H} \in \mathbb{R}^{H \times d_a}$. Our goal is to learn a policy $\pi_\theta$ that approximates the expert distribution within $\mathcal{D}_{real}$. Since learning robust policies is inherently data-hungry, relying solely on costly real-world data is impractical. Consequently, recent works have leveraged scalable simulation and human data, despite their distinct distributional shifts:

\noindent\textbf{Simulation Data ($\mathcal{D}_{sim}$).} Simulation shares the robot action parameterization and nominal kinematic model with the real world ($\mathcal{A}_{sim} \approx \mathcal{A}_{real}$). However, due to rendering artifacts and simplification, the visual observation distribution suffers from a significant \textit{sim-to-real visual gap}, i.e., $P_{sim}(o_t|s^{env}_t) \neq P_{real}(o_t|s^{env}_t)$ given the environment state $s^{env}_t$. This misalignment renders visual features learned purely from simulation unreliable for real-world deployment.

\noindent\textbf{Human Data ($\mathcal{D}_{hum}$).} Conversely, human data provides real-world visual observations whose scene statistics (lighting, textures, clutter, background) are closer to those of the real world than synthetic rendering. However, the difference in morphology introduces a fundamental \textit{human-to-robot embodiment gap}, $\mathcal{A}_{hum} \neq \mathcal{A}_{real}$, so that human actions cannot be directly executed by the robot without explicit retargeting or adaptation. In our pipeline, $\mathcal{A}_{hum}$ is a 44-dimensional vector covering both hands' wrist poses and fingertip coordinates (see Appendix~\ref{appendix:human_pose} for the exact representation); rather than performing explicit retargeting, we bridge the embodiment gap implicitly via modular encoders and decoders introduced in Section~\ref{sec:model_arch}.

\noindent\textbf{Objective.} Instead of explicitly aligning these distributions, our objective is to learn a unified policy capable of extracting complementary priors from both sources—specifically, kinematic control priors from $\mathcal{D}_{sim}$ and visual semantic priors from $\mathcal{D}_{hum}$—thereby generalizing to the target domain $\mathcal{D}_{real}$ without requiring extensive real-world data.

\begin{figure}[t]
    \centering
    \includegraphics[width=\linewidth]{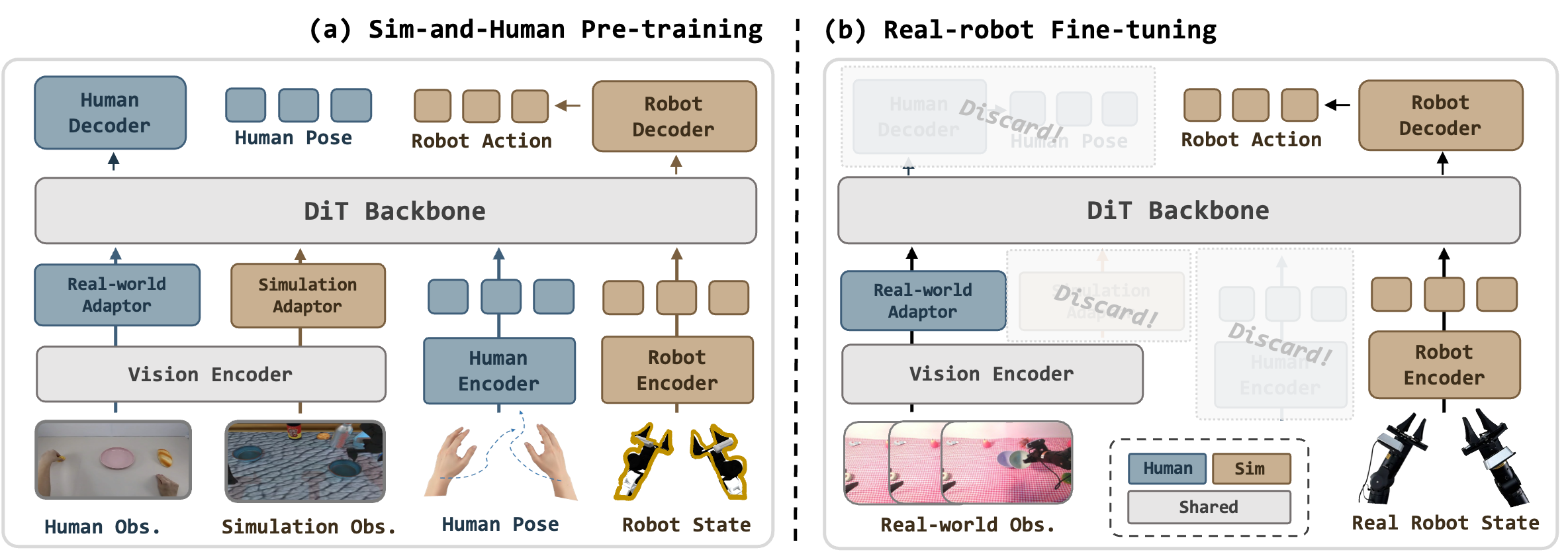}
    \caption{\textbf{Model Overview.} Our approach consists of two main stages: (1) Sim-and-Human Pre-training that routes simulation and human data through source-specific encoders/decoders so that complementary visual and kinematic priors are absorbed into the shared backbone, and (2) Real-robot fine-tuning that couples the Real-world Adaptor with the Robot Action Encoder/Decoder to enable data-efficient real-robot manipulation after limited fine-tuning.}
    \label{fig: model_arch}
\end{figure}

\subsection{Data Collection Pipeline}
\label{sec:data_pipeline}

To reduce avoidable kinematic and visual mismatch between our heterogeneous data sources and the real robot, we implement a protocol that aligns the kinematic space of $\mathcal{D}_{sim}$ and the visual space of $\mathcal{D}_{hum}$ with the target robot setup. Specifically, $\mathcal{D}_{sim}$ uses robot URDFs identical to the real robot, aligning the action parameterization and kinematic constraints so that learned action priors are more readily transferable. Similarly, $\mathcal{D}_{hum}$ captures images using the same camera model from an ego-centric viewpoint approximately aligned with the robot's head-mounted camera, encouraging transfer of scene-level visual features. Furthermore, to capture task-relevant manipulation priors, we collect data for an identical set of tasks across all three sources. Please refer to Appendix~\ref{appendix:data_collection_details} for additional details on data collection and processing.

\subsection{Modular Policy Architecture}
\label{sec:model_arch}

Our model architecture is constructed upon the DiT policy~\citep{dasari2024ditpi} where image observations are encoded by the vision encoder and the proprioceptive state is projected via a shallow MLP. The observation tokens and the noise token serve as input to an encoder-decoder transformer, which generates the denoising output. To effectively learn from both simulation and human data and extract transferable priors, we incorporate several modular components into the architecture. The overall architecture of our model is illustrated in Figure~\ref{fig: model_arch}(a).

\noindent\textbf{Modular Action Encoder and Decoder.} We utilize the human encoder to project human hand poses into latent tokens, and the corresponding decoder to translate the predicted latent actions back into the human pose space. Similarly, we employ a robot-specific encoder and decoder to process robot proprioceptive states and control actions from simulation data in the same manner. This design encourages the backbone to absorb shared task-level manipulation semantics in a unified latent space, while keeping embodiment-specific kinematic details inside the encoder and decoder.

\noindent\textbf{Domain-specific Vision Adaptors.} For both simulation and human visual observations, we first employ a shared vision encoder to extract the initial visual tokens. Subsequently, we process the simulation tokens via a Simulation Adaptor and the human tokens via a Real-world Adaptor, respectively. Structurally, both adaptors are designed as two-layer MLPs, following common practice for projecting pre-trained visual features into policy-specific latent spaces. This design routes each source through its own adaptor while sharing a common visual representation in the vision encoder, encouraging the specialization of domain-invariant features in the vision encoder and domain-specific features in the adaptors.

\subsection{Two-Stage Training Paradigm}
\label{sec:training_recipe}

\noindent\textbf{Sim-and-Human Pre-training.} In this stage, we aim to simultaneously extract the robot kinematic prior from $\mathcal{D}_{sim}$ and the real-world visual prior from $\mathcal{D}_{hum}$ into our modular policy $\pi_\theta$. We conduct joint training on both datasets, optimizing $\pi_\theta$ via the standard noise-prediction loss $\mathcal{L}(\theta; \mathcal{D})$ used in diffusion models, and combining the two sources through a weighted objective:
\begin{equation}
    \mathcal{L}_{total} = (1-\alpha)\cdot\mathcal{L}(\theta; \mathcal{D}_{sim}) + \alpha\cdot\mathcal{L}(\theta; \mathcal{D}_{hum})
\end{equation}
\noindent where $\alpha \in [0, 1]$ is the co-training ratio balancing the relative weight of simulation and human data. Following~\citep{maddukuri2025simandreal}, we implement $\alpha$ by controlling the data distribution within each mini-batch: for a batch size of $B$, we sample $\alpha \cdot B$ transitions from $\mathcal{D}_{hum}$ and $(1-\alpha) \cdot B$ from $\mathcal{D}_{sim}$, ensuring the effective gradient updates match the weighted objective.

\noindent\textbf{Real-robot Fine-tuning.} To adapt the pre-trained policy to the target robot, we reassemble its architecture as depicted in Figure~\ref{fig: model_arch}(b). We \emph{retain} four modules: the Real-world Adaptor (from the human stream, providing the real-world visual prior), the Robot Action Encoder/Decoder (from the simulation stream, providing the robot-aligned kinematic prior), the shared Vision Encoder, and the DiT Backbone; the Simulation Adaptor and the Human Action Encoder/Decoder are \emph{discarded} since their data sources are absent at fine-tuning time. The reassembled policy is then jointly fine-tuned end-to-end on a small real-robot dataset under the single learning rate listed in Appendix~\ref{appendix:hyperparameters}, yielding a data-efficient policy that leverages the complementary strengths of both pre-training sources.

\section{Experiments}
\label{sec:experiments}

\subsection{Experiment Setup}
\label{sec:experiments_setup}

\phantomsection\label{subsec:evaluation tasks}%
\textbf{Tasks.} We evaluate on four bimanual tabletop tasks that together stress different skills: \textit{Stack Bowls Two} tests sequential object interaction through fetching two bowls from separate locations and stacking them; \textit{Click Bell} tests fine-grained precision by positioning a gripper tip onto a small service-bell button; \textit{Put Bread Cabinet} tests composite long-horizon planning by grasping bread with one arm while the other opens a drawer; and \textit{Grab Roller} tests synchronized bimanual coordination by lifting a long roller horizontally with both grippers.

\phantomsection\label{subsec:Data Composition Factors}%
\textbf{Data Collection.} Following~\citep{maddukuri2025simandreal}, we decompose each dataset along five composition factors---target object ($\mathcal{F}_{obj}$), visual distractors ($\mathcal{F}_{dist}$), lighting ($\mathcal{F}_{light}$), background ($\mathcal{F}_{bg}$), and initial pose ($\mathcal{F}_{init}$)---that jointly parameterize the training distribution and guide both data collection and evaluation. \textbf{For simulation,} we use RoboTwin2.0~\citep{chen2025robotwin} with its default domain randomization to generate 500 kinematically feasible trajectories per task. \textbf{For human data,} we collect 500 demonstrations per task across 12 scenarios, so that the model inherits a real-world visual prior from these environments. \textbf{For real-robot data,} we deliberately constrain each task to 80 episodes---50 in a canonical base scene plus 30 across three complex scenes (10 each)---on which all methods are subsequently fine-tuned for a fair comparison. Notably, although the simulation and human datasets are $\mathbf{6\times}$ larger than the real-robot set, both pipelines collect at $\mathbf{5\times}$ real-robot throughput (per-task breakdown in Appendix~\ref{appendix:collection_throughput}), making the total collection effort comparable across all three sources.

\phantomsection\label{subsec:evaluation settings}%
\textbf{Evaluation Protocol.} We report Success Rate (SR) and Progress Rate (PR), measuring final task completion and the completion percentage of sub-goals respectively. Each policy is evaluated under two settings. The \textit{in-distribution} (ID) setting comprises one canonical base scenario (clean background, stable lighting) and three complex scenarios with diverse textures, moderate distractors, and dynamic lighting---10 randomized trials per scenario, 40 trials in total. The \textit{out-of-distribution} (OOD) setting features two extreme-complex scenarios assembled from held-out backgrounds, distractors, object instances, and lighting that never appeared in the human or real-robot datasets used for pre-training and fine-tuning, with 10 trials each (20 trials total). These factors also lie outside the simulation randomization, which we discuss in Appendix~\ref{appendix:env_protocols}. Each configuration is a single trained policy, and $\pm$ denotes the standard error over its evaluation trials (trial counts per experiment and a second-seed check in Appendix~\ref{appendix:trials_seeds}).

\textbf{Baselines.} We compare \textbf{SimHum}---our full method pre-trained on both simulation and human data---against three configurations that isolate each prior. \textit{Real only} is trained from scratch on the real-robot dataset and serves as the no-pre-training reference. \textit{HumReal} pre-trains on human demonstrations alone, so its robot action encoder/decoder receives no pre-training signal and starts fine-tuning from random initialization, isolating the contribution of the real-world visual prior. \textit{SimReal} pre-trains on simulation data alone and fine-tunes with its simulation-pre-trained robot action encoder/decoder and its Simulation Adaptor in place of the Real-world Adaptor, isolating the contribution of the robot kinematic prior. All four variants share the same architecture, training hyperparameters, and fine-tuning real-robot data, so performance differences are attributable to the pre-training source mixture, which determines both what the shared modules learn and which modules arrive pre-trained at fine-tuning.

Per-task scene configurations, factor definitions, evaluation scene layouts, and training hyperparameters are deferred to Appendix~\ref{appendix:eval_setup}--\ref{appendix:hyperparameters}.

\begin{figure}[t]
    \centering
    \includegraphics[width=\linewidth]{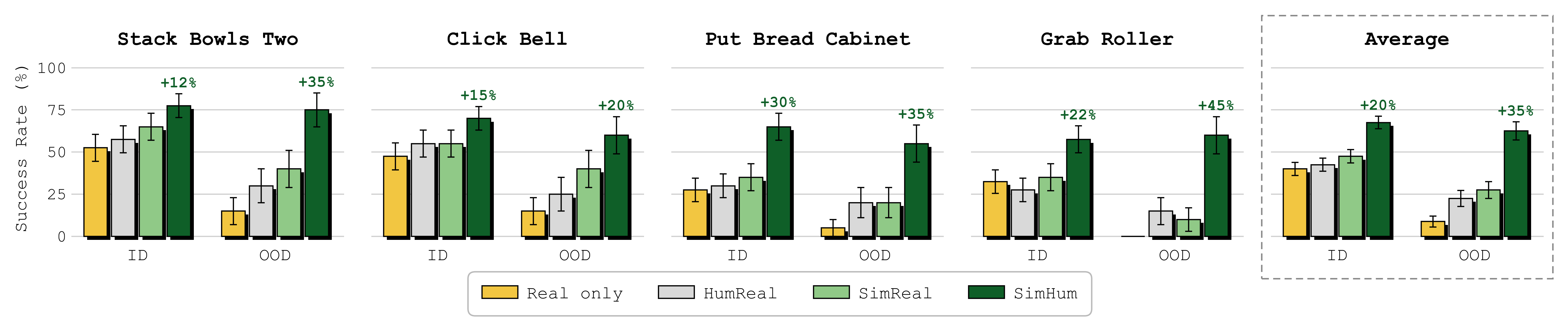}
    \caption{\textbf{SimHum achieves the best ID/OOD performance.} We report per-task and average success rates under ID and OOD settings. More detailed results in Appendix~\ref{appendix:full_main_results}.}
    \label{fig:main_results}
\end{figure}

\subsection{Effectiveness of Sim-and-Human Co-training}
\label{subsec:main_results}

\textbf{SimHum enables robust policy generalization in novel scenes.} We evaluate all methods on four real-world bimanual tasks under ID and OOD settings (Figure~\ref{fig:main_results}), from which we draw three conclusions. First, Real only policies perform reasonably under ID but degrade sharply under OOD, suggesting sensitivity to scene changes under limited real-data fine-tuning. Second, single-source pre-training (SimReal and HumReal) brings only bounded improvements, because each is bottlenecked by a single domain gap---visual for simulation, embodiment for human data---which the limited real-robot data cannot fully close. Finally, SimHum extracts complementary priors from both sources and consistently outperforms all baselines across tasks. Most notably under OOD, it surpasses the strongest single-source baseline by $\mathbf{35.0\%}$ and the Real only policy by $\mathbf{53.7\%}$ in absolute average SR, demonstrating robust generalization to held-out novel scenes.

\subsection{Decoupling the Effects of Simulation and Human data}
\label{subsec: data contributions analysis}

\begin{figure}[!t]
    \centering
    \includegraphics[width=\linewidth]{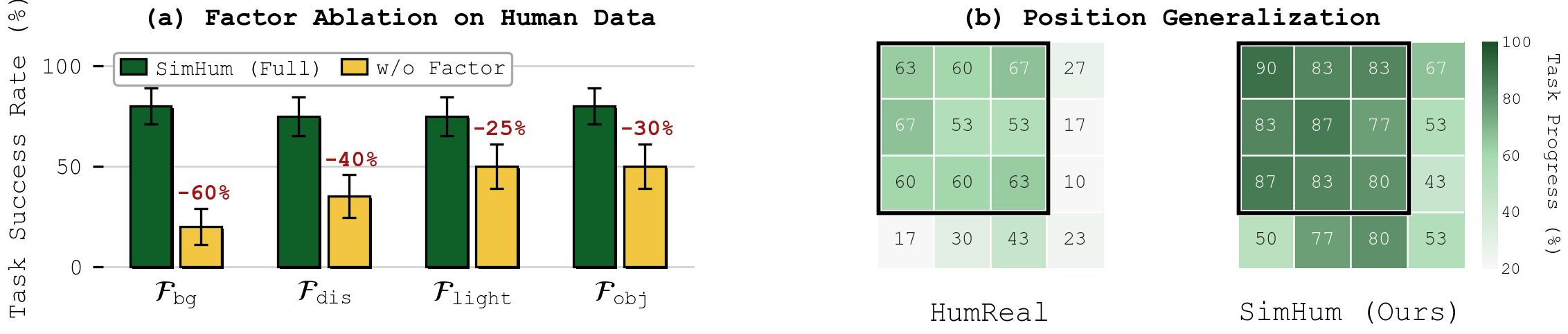}
    \caption{\textbf{Simulation and human data play distinct roles.} \textbf{(a)} Leave-one-factor-out ablation on human data, where each policy is evaluated on the OOD scenario probing the removed factor. Removing any factor lowers OOD SR, most for background. \textbf{(b)} Per-position PR on a $4\times4$ grid in an OOD scene. Human and real-robot demonstrations cover only the inner $3\times3$ cells (outlined in black), whereas simulation samples positions continuously over a broader range. HumReal degrades most at the unseen outer cells.}
    \label{fig:decoupling}
\end{figure}

\textbf{Human data improves visual OOD generalization.} To isolate the contribution of human data, we conduct a leave-one-factor-out ablation over $\{\mathcal{F}_{obj}, \mathcal{F}_{bg}, \mathcal{F}_{light}, \mathcal{F}_{dist}\}$: for each factor, we remove its diversity from the human dataset and evaluate the resulting policy on an OOD scenario that specifically probes that factor. Appendix~\ref{appendix:human_data_ablation_details} illustrates the construction using the background factor as a representative example. As shown in Figure~\ref{fig:decoupling}(a), removing any single factor decreases the OOD success rate, and removing background diversity has the largest effect, lowering SR from 80.0\% to 20.0\%, consistent with its role as the dominant visual perturbation in robot observations. This consistent pattern across all four factors suggests that the diversity in human data equips the model with a visual prior that transfers to real-world scenes.

\textbf{Simulation data enhances policy robustness to novel object positions.} We fine-tuned SimHum and HumReal on 80 real-robot episodes, with initial object positions in both the real-robot and human demonstrations confined to a fixed $3 \times 3$ grid. Simulation, in contrast, samples object positions continuously over a broader range. For evaluation, objects were positioned across an expanded $4 \times 4$ grid within an OOD scenario, and we report the average Progress Rate (PR) per grid location aggregated over 10 trials. Figure~\ref{fig:decoupling}(b) shows that HumReal is competent within seen positions (60.7\% PR on average) but struggles with spatial generalization, suffering a $\mathbf{-36.9\%}$ decay in absolute PR in unseen regions. In contrast, SimHum leverages diverse robot action trajectories from simulation and improves seen-zone performance by $\mathbf{+23\%}$ while extending generalization to unseen areas by $\mathbf{+36.7\%}$ over HumReal in absolute PR. This supports the interpretation that simulation's broad coverage of object positions provides kinematic priors that transfer to novel positions.

\begin{figure}[!t]
    \centering
    \includegraphics[width=\linewidth]{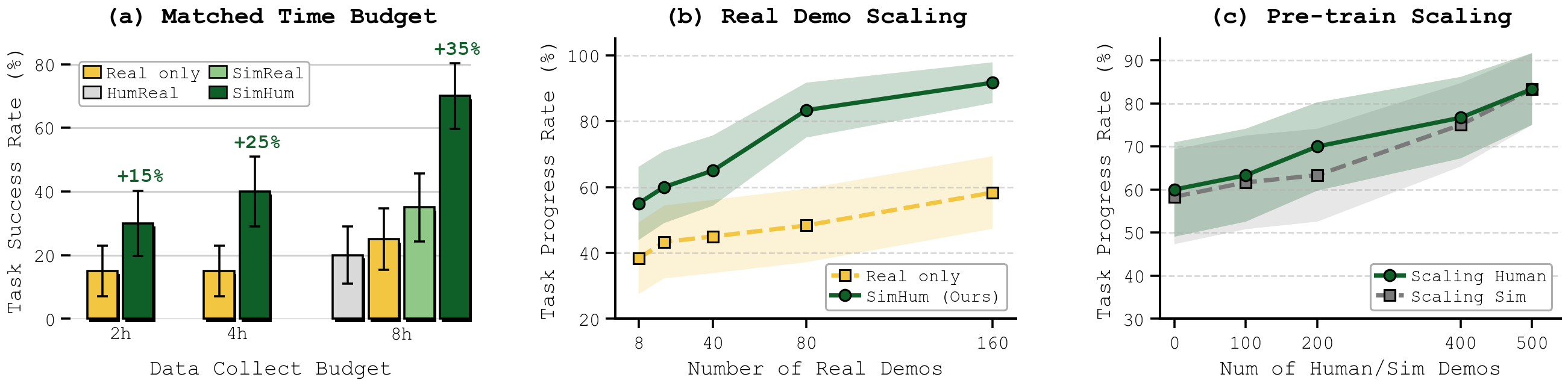}
    \caption{\textbf{(a) Matched collection-time budgets.} Under 2/4/8\,h budgets, SimHum splits time between real-robot teleoperation and human collection, and simulation is generated in parallel without being charged against any budget. SimHum outperforms Real only at every budget and both single-source baselines at 8\,h. \textbf{(b) Scaling real-robot demonstrations.} With sim+human pre-training, SimHum consistently outperforms Real only across all real-robot data scales (8 to 160 demos), demonstrating the value of pre-training. \textbf{(c) Scaling pre-training sources.} With fixed real-robot data, increasing either pre-training source consistently improves SimHum. All experiments are conducted on Stack Bowls Two under OOD; shading denotes standard error.}
    \label{fig:scaling}
\end{figure}

\FloatBarrier

\subsection{Data Efficiency and Performance Scalability}
\label{subsec:scaling}

To probe data efficiency and scaling under a controlled setting, we conduct three studies on \textit{Stack Bowls Two} under OOD, sweeping different axes of the data composition (Figure~\ref{fig:scaling}).

\textbf{SimHum achieves better data efficiency under limited collection budgets.} Across 2/4/8\,h total budgets, SimHum splits time evenly between real-robot teleoperation and human collection, with simulation generated in parallel and not charged against any method's budget (Figure~\ref{fig:scaling}(a); configurations in Appendix~\ref{appendix:time_budget_details}). SimHum outperforms Real only at every budget, by up to $\mathbf{45.0\%}$ in absolute SR at 8\,h. At 8\,h it also surpasses SimReal by $\mathbf{35.0\%}$ and HumReal by $\mathbf{50.0\%}$ in absolute SR, suggesting that combining the two sources yields gains beyond either source on its own.

\textbf{Policy performance scales with real-robot dataset size.} To probe how OOD performance scales with the size of the real-robot fine-tuning set, we vary the number of demonstrations from 8 to 160 (Figure~\ref{fig:scaling}(b)). SimHum outperforms Real only at every scale. With only 8 demos, SimHum already reaches the level of Real only trained on 160 demos in both PR and SR, with the full SR results reported in Appendix~\ref{appendix:demo_scaling_sr}. This indicates that pre-training on simulation and human data yields priors useful for OOD generalization and reduces reliance on real-robot data.

\textbf{Policy performance scales with pre-training data.} To isolate the contribution of each pre-training source, we adopt a controlled protocol: hold one source fixed at 500 episodes while sweeping the other from 0 to 500 (Figure~\ref{fig:scaling}(c)). \textit{Either} axis---simulation or human---yields consistent gains in PR, suggesting that the two priors contribute independently. Moreover, the curves are still rising at 500 episodes, suggesting that continued scaling of either source has yet to saturate and may yield further gains with more data.

\subsection{Ablation Study on SimHum}
\label{subsec:ablation}

\begin{wraptable}[12]{r}{0.49\linewidth}
\vspace{-12pt}
\centering
\caption{\textbf{Ablations on SimHum.} All experiments are conducted on Stack Bowls Two under OOD. Removing any component or the hand-pose supervision lowers SR.}
\label{tab:component_ablation}
\small
\setlength{\tabcolsep}{5pt}
{%
\begin{tabular}{lc}
\toprule
\textbf{Method} & {\textbf{SR} $\uparrow$} \\
\midrule
SimHum w/o vision adaptors    & \acc{40.0}{11.0} \\
SimHum w/o real-world adaptor & \acc{55.0}{11.1} \\
SimHum w/o relative action    & \acc{45.0}{11.1} \\
SimHum w/ visual-only human loss & \acc{40.0}{11.0} \\
\rowcolor{lightcyan}
\textbf{SimHum (Ours)} & \bfseries \acc{75.0}{\phantom{1}9.7} \\
\bottomrule
\end{tabular}}
\vspace{-8pt}
\end{wraptable}

\begin{wrapfigure}[15]{r}{0.5\linewidth}
\vspace{-4pt}
\centering
\includegraphics[width=\linewidth]{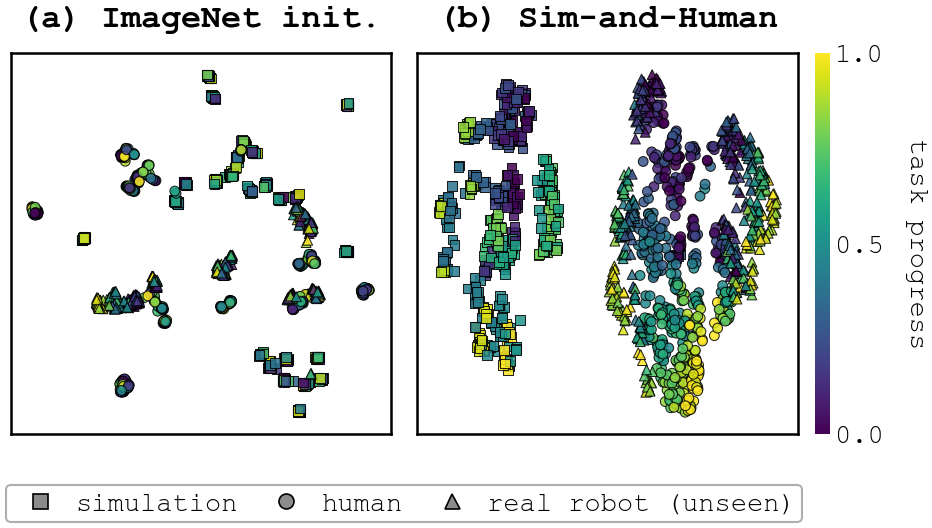}
\caption{\textbf{Representation visualization by UMAP~\citep{mcinnes2018umap} on Stack Bowls Two.} Frozen vision-encoder features, with shape denoting the data source and color denoting task progress. After pre-training, frames from all three sources are ordered by task progress.}
\label{fig:representation_umap}
\vspace{-10pt}
\end{wrapfigure}

\textbf{Impact of SimHum Components.} In Table~\ref{tab:component_ablation}, we report OOD SR for four ablations on Stack Bowls Two and find that every component is necessary. \textbf{First,} removing both vision adaptors causes the largest drop, lowering SR to 40.0\%, indicating that domain-specific routing separates simulation and human visual features. \textbf{Second,} using the Simulation Adaptor instead of the Real-world Adaptor during fine-tuning lowers SR to 55.0\%, indicating that the Real-world Adaptor preserves the real-world visual prior. \textbf{Third,} using absolute rather than relative actions lowers SR to 45.0\%, indicating that relative actions help unify heterogeneous coordinate systems. \textbf{Finally,} replacing the hand-pose prediction loss on human data with an R3M-style visual-only objective~\citep{nair2022r3m} lowers SR to 40.0\%. This indicates that task-relevant hand-pose supervision is more helpful than visual-only supervision for learning a transferable real-world visual prior. Our representation visualization (Figure~\ref{fig:representation_umap}) is consistent with this point.

\begin{wrapfigure}[10]{r}{0.42\linewidth}
\vspace{-16pt}
\centering
\includegraphics[width=0.74\linewidth]{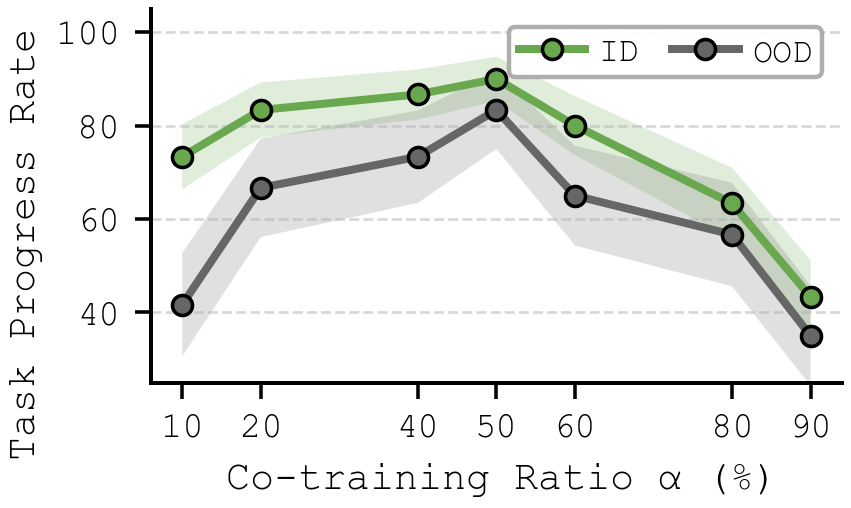}
\caption{\textbf{Co-training ratio ablation.} Evaluated on Stack Bowls Two. Performance is highest near $\alpha=50\%$ and stable across balanced mixtures.}
\label{fig:alpha_ablation}
\vspace{-8pt}
\end{wrapfigure}
\textbf{Impact of Co-training ratio.} We study the impact of the co-training ratio $\alpha$, which controls the proportion of human data per batch. Figure~\ref{fig:alpha_ablation} shows that performance is highest near $\alpha=50\%$ and stable across balanced mixtures, suggesting that both sources matter. As $\alpha$ rises to $90\%$, both ID and OOD performance decline because the dominance of human data weakens the kinematic prior from simulation, biasing the policy toward human-specific motion patterns that are difficult to transfer to robot control. Conversely, at $\alpha=10\%$, OOD degrades far more sharply than ID because the dominance of simulation data weakens the visual prior from human data, leaving the policy unable to generalize to novel scenes; ID performance is less affected since real-robot fine-tuning provides direct supervision on ID scenes.

\needspace{16\baselineskip}
\textbf{What does the shared representation learn?} To understand why priors learned from simulation and human data transfer to the real robot, we visualize the visual representation learned by the shared vision encoder. On Stack Bowls Two, we sample 25 episodes with 20 frames each from simulation, human, and real-robot data, where the real-robot frames were never seen in pre-training, and project their frozen features to 2D with UMAP. As shown in Figure~\ref{fig:representation_umap}(a), under the ImageNet initialization the frames collapse into per-episode clusters with no progress structure. In contrast, after Sim-and-Human pre-training (Figure~\ref{fig:representation_umap}(b)), frames from all three sources, including the unseen real-robot frames, are ordered by task progress. This suggests that the shared representation captures task progress rather than scene appearance alone, which may explain why the visual prior from human data transfers to unseen real-robot scenes.

\section{Conclusion}
\label{sec:conclusion}

In this work, we present \textbf{SimHum}, a co-training recipe for data-efficient bimanual manipulation that leverages the complementary strengths of simulation and human demonstrations. SimHum uses simulation to provide robot-aligned kinematic priors and human data to provide real-world visual priors, then adapts these priors with a small real-robot dataset. Across four bimanual tabletop tasks, SimHum achieves strong OOD generalization: with only 80 real-robot episodes per task, it attains a 62.5\% OOD success rate, 53.7\% higher than Real only in absolute success rate. Moreover, in a controlled data-collection study with matched collection time, SimHum improves over the best single-source pre-training baseline by 35.0\% in absolute success rate, suggesting that co-training on simulation and human data can further reduce real-robot data requirements beyond single-source pre-training.

\noindent\textbf{Limitations and Future Work.} We highlight four directions. \textbf{First,} our evaluation focuses on real-world bimanual tabletop manipulation and does not cover the full failure space of manipulation in the wild; broader studies across dexterous in-hand manipulation, long-horizon multi-stage tasks, and humanoid or mobile manipulators are needed to characterize where the recipe transfers. \textbf{Second,} our current setup focuses on task-aligned policy learning where simulation, human, and real-robot data share the same task set. Applying the Sim-and-Human recipe to large-scale foundation-policy pre-training, such as a Vision-Language-Action backbone, would involve multi-task, language-conditioned training at a much larger scale, which we leave to future work. \textbf{Third,} our human data is task-aligned and self-collected, captured from tripod-mounted views aligned with the robot's head camera. Scaling SimHum to unconstrained egocentric data such as Ego4D~\citep{grauman2022ego4d} or EgoExo4D~\citep{grauman2024egoexo4d} remains open. \textbf{Finally,} the remaining ID/OOD reliability gap is concentrated at the execution layer---bimanual coordination and precise contact timing---rather than perception, motivating closed-loop contact feedback and residual reinforcement learning on top of SimHum's priors as natural next steps.

\clearpage
\acknowledgments{This study is supported by grants from the National Natural Science Foundation of China (\mbox{No.~U23A20315}, \mbox{No.~U22A2097}), and Sichuan Science and Technology Program (\mbox{No.~2025ZNSFSC1493}).}

\bibliography{references}

\clearpage
\noindent{\Large\bfseries Appendix}\par
\vspace{1em}
\appendix
\section{Appendix Overview}
\label{sec:appendix_overview}

This appendix is organized as follows:
\begin{itemize}
    \item \textbf{Tasks, Metrics, and Evaluation Protocol} (Appendix~\ref{appendix:eval_setup}): defines the four manipulation tasks with milestone-based scoring, the SR/PR metrics, the in-distribution / out-of-distribution evaluation protocol, and the trial and seed counts.
    \item \textbf{Data Collection} (Appendix~\ref{appendix:data_collection_details}): hardware systems and operator setups, environmental protocols and scenarios, action representation and processing, and per-source throughput.
    \item \textbf{Implementation Details and Hyperparameters} (Appendix~\ref{appendix:hyperparameters}): visual encoder, diffusion policy backbone, training hyperparameters, and training scope.
    \item \textbf{Protocols for Decoupling and Scaling Experiments} (Appendix~\ref{appendix:exp_setup}): protocols for the human-data factor ablation, position generalization analysis, and time-budget scaling configurations.
    \item \textbf{Additional Results} (Appendix~\ref{appendix:additional_results}): full per-task SR and PR, success rates for scaling real-robot demonstrations, additional comparisons, and cross-source behavior mismatch.
\end{itemize}

\section{Tasks, Metrics, and Evaluation Protocol}
\label{appendix:eval_setup}

This appendix groups the task setup used throughout our experiments: task definitions and milestone-based scoring (\S\ref{appendix:tasks}), success and progress metrics (\S\ref{appendix:eval_metrics}), and the in-distribution / out-of-distribution evaluation protocol (\S\ref{appendix:eval_settings}).

\subsection{Task Definitions and Scoring Criteria}
\label{appendix:tasks}

We evaluate \textbf{SimHum} on four bimanual tabletop tasks: Stack Bowls Two, Click Bell, Put Bread Cabinet, and Grab Roller. Figure~\ref{fig:tasks} visualizes the task stages and milestone scores. The milestone scoring lets us report both binary task success and partial progress, separating complete failures from executions that finish only early sub-goals. Detailed task descriptions and scoring criteria are listed below:

\begin{figure}[t]
    \centering
    \includegraphics[width=\linewidth]{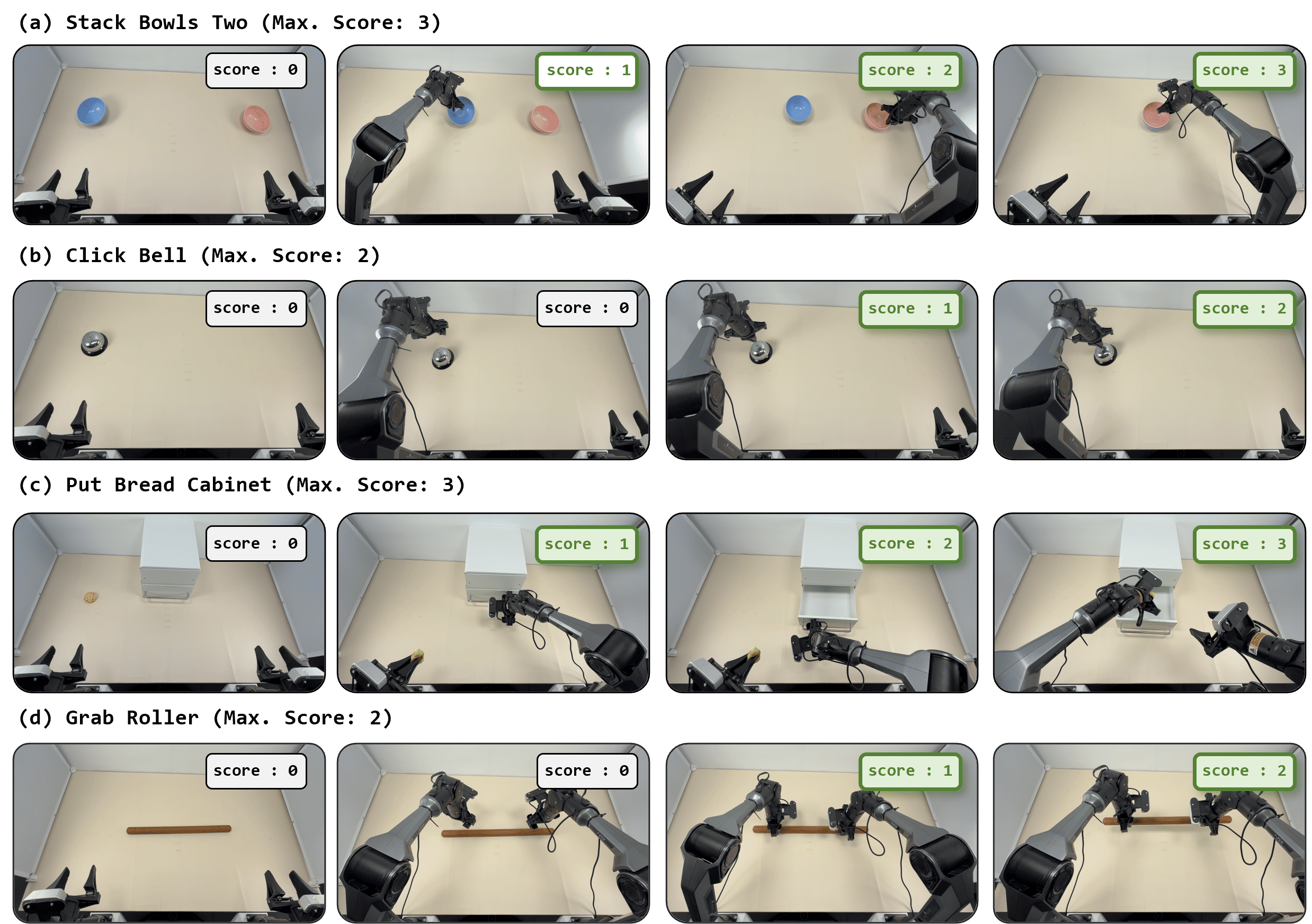}
    \captionsetup{width=0.84\linewidth}
    \caption{\textbf{Task Definitions and Scoring.} Visual breakdown of the four manipulation tasks (Stack Bowls Two, Click Bell, Put Bread Cabinet, Grab Roller) and their respective success/progress criteria.}
    \label{fig:tasks}
\end{figure}

\noindent\textbf{Stack Bowls Two (Max Score: 3).} This task requires precise dual-arm coordination to stack two bowls initially placed apart. The scoring reflects the cumulative success of independent arm actions and their coordination.
\begin{itemize}
    \item \textbf{0:} No meaningful interaction with either bowl, or failure to grasp.
    \item \textbf{1:} Successfully grasping the left bowl with the left arm.
    \item \textbf{2:} Successfully grasping both bowls (one in each hand).
    \item \textbf{3:} Precisely stacking one bowl into the other without dropping either.
\end{itemize}

\noindent\textbf{Click Bell (Max Score: 2).} This task assesses the agent's fine motor control and precision in actuating a small target.
\begin{itemize}
    \item \textbf{0:} Failure to reach the target area.
    \item \textbf{1:} Moving the end-effector to directly above the bell.
    \item \textbf{2:} Applying downward force to depress the button and produce a ringing sound.
\end{itemize}

\noindent\textbf{Put Bread Cabinet (Max Score: 3).} This is a long-horizon task involving articulated object manipulation and multi-stage planning.
\begin{itemize}
    \item \textbf{0:} Failure to grasp the bread object from the table.
    \item \textbf{1:} Successfully grasping the bread object.
    \item \textbf{2:} Successfully pulling the cabinet door open.
    \item \textbf{3:} Placing the bread inside the cabinet and releasing it.
\end{itemize}

\noindent\textbf{Grab Roller (Max Score: 2).} This task evaluates bimanual synchronization and grasp stability on cylindrical objects.
\begin{itemize}
    \item \textbf{0:} Failure to establish a stable grasp on the roller.
    \item \textbf{1:} Successfully establishing a grasp on the roller stick with both grippers.
    \item \textbf{2:} Lifting the roller to the target height with both arms while maintaining grasp without the object falling.
\end{itemize}

\subsection{Evaluation Metrics}
\label{appendix:eval_metrics}

To measure both the final task completion and the quality of partial execution, we employ the following metrics. The specific milestone definitions and maximum scores ($S_{max}$) for each task are detailed in Appendix~\ref{appendix:tasks}.

\noindent\textbf{Success Rate (SR).} SR measures the percentage of trials where the policy achieves the maximum possible score (i.e., complete task success). It is calculated as:
\begin{equation}
    SR = \frac{1}{N} \sum_{i=1}^{N} \mathbb{I}(S_i = S_{max}) \times 100\%
\end{equation}
where $N$ is the total number of evaluation trials, $S_i$ represents the score achieved in the $i$-th trial, $S_{max}$ is the maximum score defined for the task, and $\mathbb{I}(\cdot)$ is the indicator function.

\noindent\textbf{Progress Rate (PR).} PR measures the average fraction of task milestones completed across trials, distinguishing complete failures from partial executions. It is defined as the ratio of accumulated score to maximum possible score:
\begin{equation}
    PR = \frac{\sum_{i=1}^{N} S_i}{N \times S_{max}} \times 100\%
\end{equation}
A higher PR indicates a larger fraction of completed milestones even without final success.

\begin{figure}[t]
    \centering
    \includegraphics[width=\linewidth]{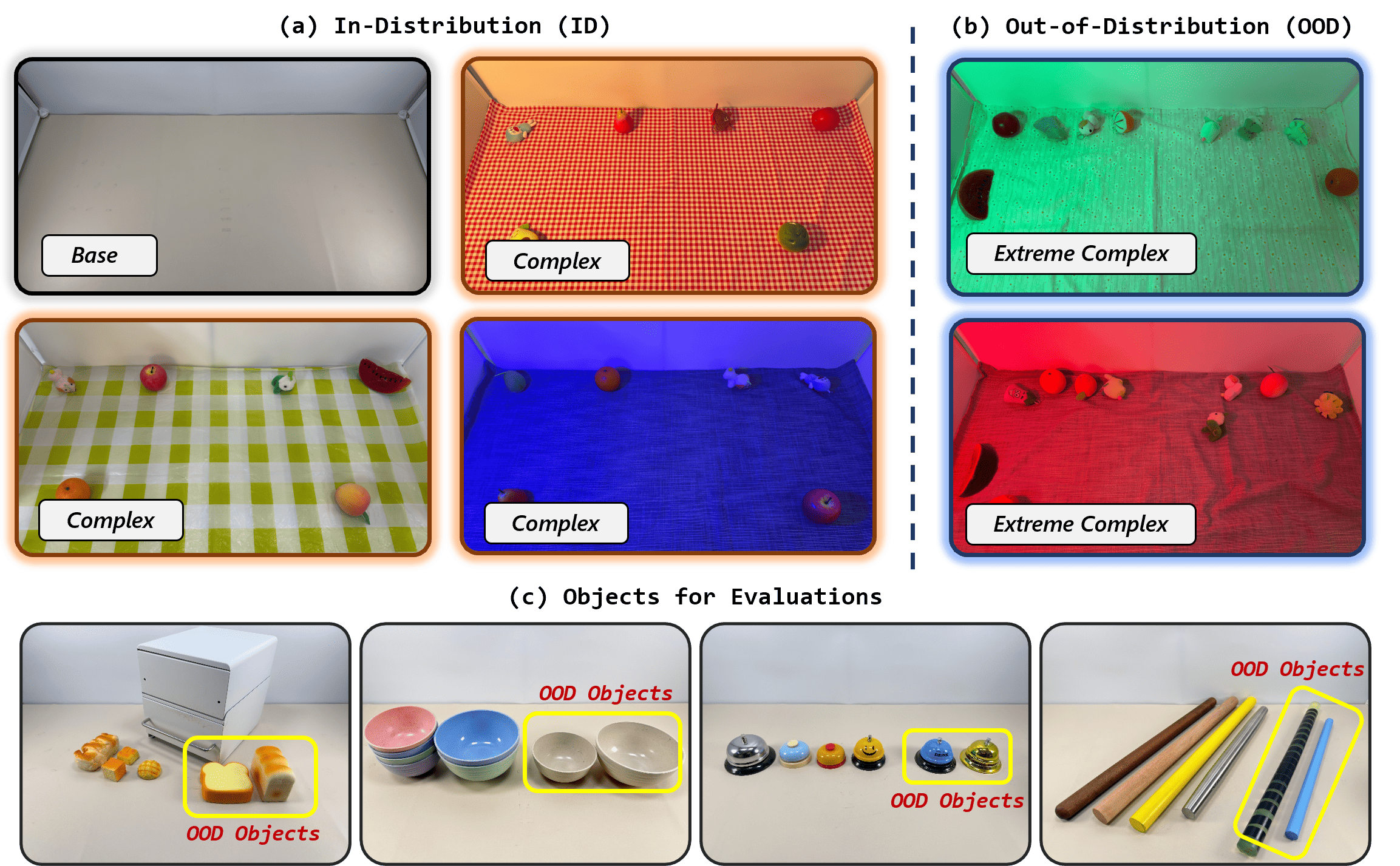}
    \caption{\textbf{Evaluation Settings.} (a) \textbf{In-Distribution (ID)} scenarios comprising a standard \textit{Base} environment and \textit{Complex} environments seen during training. (b) \textbf{Out-of-Distribution (OOD)} scenarios labeled as \textit{Extreme Complex}, constructed using strictly held-out environmental factors (e.g., unseen lighting, novel textures). (c) \textbf{Evaluation Objects} across tasks, where yellow bounding boxes indicate specific held-out instances (OOD Objects) introduced to test generalization to OOD scenes.}
    \label{fig:eval_setting}
\end{figure}

\subsection{Evaluation Settings}
\label{appendix:eval_settings}

We classify the evaluation scenarios into In-Distribution (ID) and Out-of-Distribution (OOD) settings based on the environmental factors defined in Section~\ref{subsec:Data Composition Factors}. Visual examples of these settings are provided in Figure~\ref{fig:eval_setting}. For each evaluation trial, objects are randomly initialized within the spatial ranges defined in Figure~\ref{fig:init_position_setting}, using the same placement ranges as data collection.

\noindent\textbf{In-Distribution (ID).} This setting evaluates the policy's performance in environments seen during the real-robot fine-tuning phase. As illustrated in Figure~\ref{fig:eval_setting}(a), it consists of two categories:
\begin{itemize}
    \item \textbf{Base Scenario (1 scenario):} A standard environment characterized by a clean white background and stable, neutral lighting conditions, serving as the baseline for robot operation.
    \item \textbf{Complex Scenarios (3 scenarios):} These scenarios introduce domain variations encountered during training. They feature backgrounds with diverse textures, a moderate number of distractor objects, and dynamic interference such as flashing blue lights.
\end{itemize}
Regarding the objects, the target items manipulated in the ID setting correspond strictly to the object instances seen during training, as depicted in the left panel of Figure~\ref{fig:eval_setting}(c). All four ID scenarios are included in the real-robot fine-tuning dataset. For each task, we conduct 10 evaluation trials per scenario, resulting in a total of 40 trials. We report the aggregated average SR and PR across these 40 trials.

\noindent\textbf{Out-of-Distribution (OOD).} To test OOD generalization, we construct two novel scenarios labeled as ``Extreme Complex'' (Figure~\ref{fig:eval_setting}(b)). In contrast to the ID settings, these scenarios are constructed from \textbf{held-out} environmental factors that do not appear in the human or real-robot training datasets. Specifically, we introduce novel background textures (featuring irregular wrinkles to increase complexity), unseen distractor objects arranged in higher density to create heavy clutter, and held-out task object instances (as highlighted in Figure~\ref{fig:eval_setting}(c)). We perform 10 evaluation trials for each of the two OOD scenarios, totaling 20 trials, and report the average SR and PR.

\subsection{Trials and Seeds}
\label{appendix:trials_seeds}

\begin{wraptable}[8]{r}{0.38\linewidth}
\vspace{-14pt}
\centering
\caption{\textbf{Component ablation over 40 OOD trials.} Stack Bowls Two, same policies as Table~\ref{tab:component_ablation}.}
\label{tab:component_ablation_n40}
\small
\setlength{\tabcolsep}{3pt}
\resizebox{\linewidth}{!}{%
\begin{tabular}{lc}
\toprule
\textbf{Method} & \textbf{SR} $\uparrow$ \\
\midrule
SimHum w/o vision adaptors    & \acc{32.5}{7.4} \\
SimHum w/o real-world adaptor & \acc{57.5}{7.8} \\
SimHum w/o relative action    & \acc{52.5}{7.9} \\
\textbf{SimHum (Ours)} & \acc{80.0}{6.3} \\
\bottomrule
\end{tabular}}
\vspace{-10pt}
\end{wraptable}
All experiments follow the evaluation protocol in Section~\ref{subsec:evaluation settings}, with 40 ID trials and 20 OOD trials from one trained policy per configuration, and the ``Average'' entries in Figure~\ref{fig:main_results} and Table~\ref{tab:main_results_full} pool the four per-task policies. Two additional checks confirm that the conclusions are not driven by trial or seed noise. First, we evaluated each of the three component ablations in Table~\ref{tab:component_ablation} for another 20 OOD trials, and the results over all 40 trials (Table~\ref{tab:component_ablation_n40}) preserve the original ordering. Second, retraining SimHum on Stack Bowls Two with a different random seed gives 80.0$\pm$8.9 OOD SR, compared with 75.0$\pm$9.7 for the seed reported in Table~\ref{tab:component_ablation}.

\section{Data Collection}
\label{appendix:data_collection_details}

This appendix details the three-source data-collection pipeline complementing Section~\ref{sec:data_pipeline}: hardware systems and operator setups (\S\ref{appendix:data_systems}), environmental protocols and scenarios (\S\ref{appendix:env_protocols}), action representation and processing (\S\ref{appendix:action_space}), and per-source throughput (\S\ref{appendix:collection_throughput}).

\subsection{Data Collection Systems}
\label{appendix:data_systems}

\begin{figure}[!ht]
    \centering
    \includegraphics[width=0.92\linewidth,trim=18pt 307pt 18pt 52pt,clip]{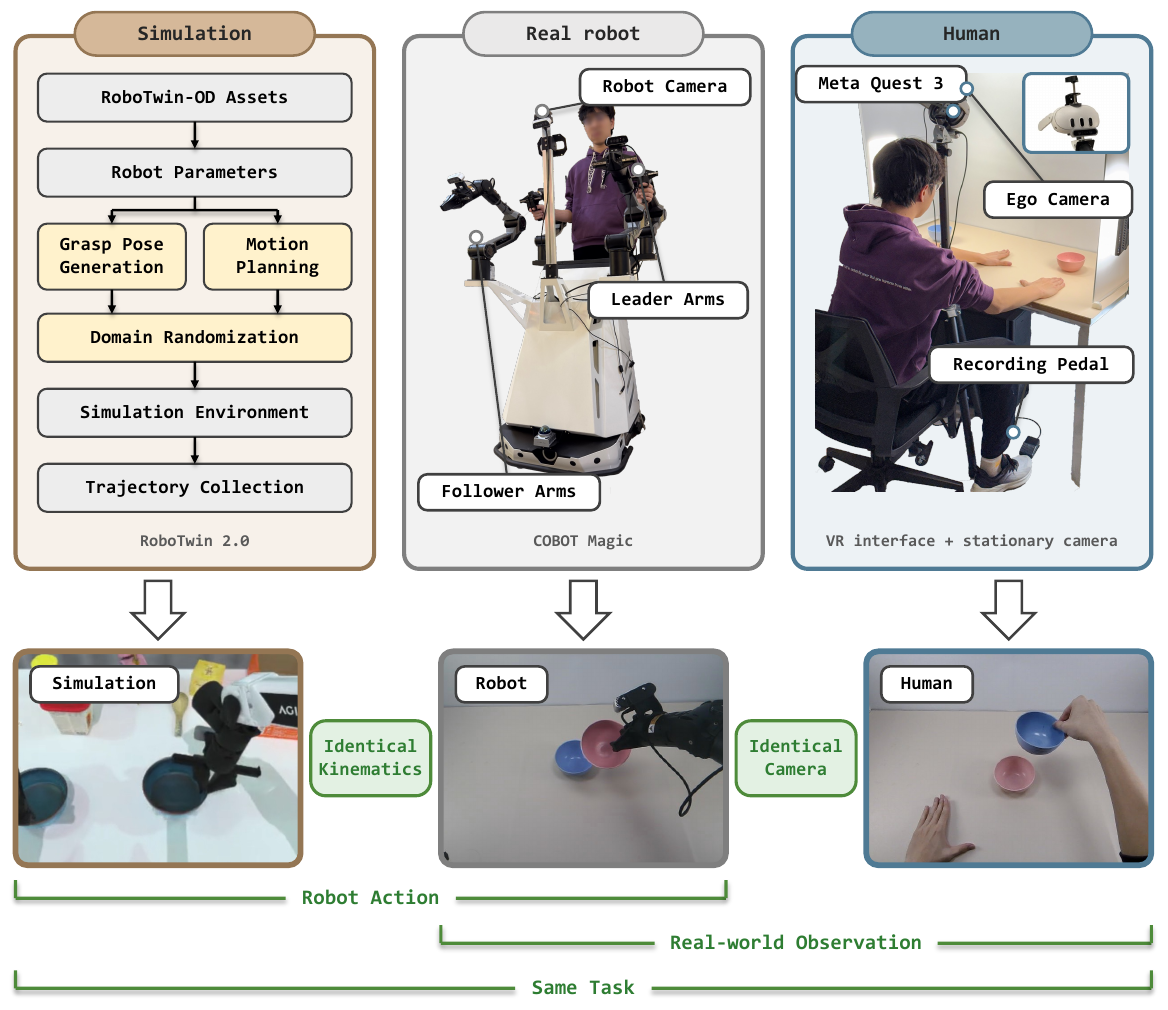}
    \caption{\textbf{Overview of Data Collection Systems.} (Left) Automated simulation data generation based on RoboTwin2.0~\citep{chen2025robotwin}; (Middle) real-robot teleoperation on COBOT Magic with a leader-follower system; (Right) human data collection with a VR interface and a stationary camera. Simulation shares identical kinematics with the real robot, and human data shares an identical camera, so all three sources cover the same tasks with complementary priors.}
    \label{fig:data_collection}
\end{figure}

We utilize three distinct pipelines to acquire manipulation data, implementing a protocol that aligns the kinematic space of $\mathcal{D}_{sim}$ and the visual space of $\mathcal{D}_{hum}$ with the target robot to reduce avoidable kinematic and visual mismatch, as illustrated in Figure~\ref{fig:data_collection}:
\begin{itemize}
    \item \textbf{Simulation Data Generation:} We use RoboTwin2.0~\citep{chen2025robotwin} to generate simulation robot trajectories. We instantiate robot assets that share identical kinematics with the physical hardware and execute tasks aligned with real-world specifications, yielding kinematically feasible trajectories that provide robot-aligned action priors.
    \item \textbf{Real-Robot Data Collection:} For physical robot demonstrations, we utilize the COBOT Magic platform. Data is collected using a \textbf{leader-follower teleoperation} setup, where the operator controls the leader arms to drive the follower arms.
    \item \textbf{Human Data Collection (Right Panel):} To capture diverse human demonstrations, we designed a cost-effective ($<\$500$) system aligned with the robot's observation space. We mount an RGB camera (identical to the robot's head-mounted sensor) and a Meta Quest 3 device together on a single stationary tripod, positioned at a height approximately aligned with the robot's head camera to preserve an ego-centric viewpoint. The camera captures visual observations while the Meta Quest 3 captures human hand poses via its built-in markerless tracker. A recording pedal provides start/stop control, reducing hand-held capture fatigue and camera jitter while preserving the first-person perspective; viewpoint and height are matched to the robot's by visual inspection, without precise calibration. Prior robotic characterization reports $\sim$1.7\,cm positional error and $\sim$1.1\,cm jitter for the Quest~3~\citep{godden2025quest}; because the Human Action Encoder/Decoder are \emph{discarded} during real-robot fine-tuning (Section~\ref{sec:training_recipe}), this tracking noise affects only the human action stream during pre-training and has limited direct impact on the fine-tuned real-robot policy.
\end{itemize}

\subsection{Environmental Protocols and Scenarios}
\label{appendix:env_protocols}

\begin{figure}[t]
    \centering
    \includegraphics[width=\linewidth]{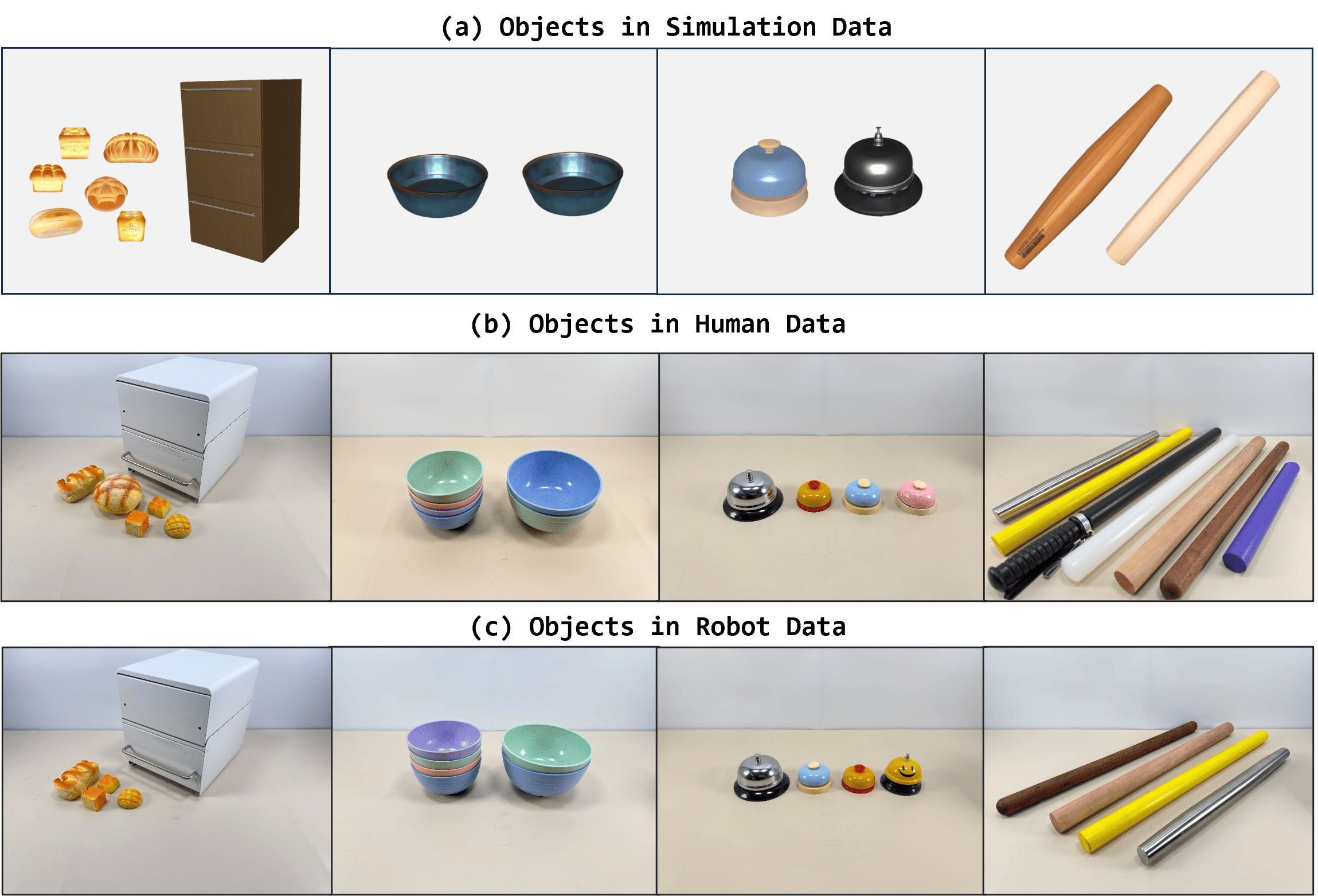}
    \caption{\textbf{Objects used in data collection.} Visualization of the specific object instances ($\mathcal{F}_{obj}$) utilized across different tasks. (a) Synthetic assets from the simulation dataset. (b) Real-world objects from the human dataset. (c) Real-world objects from the real-robot dataset.}
    \label{fig:objects}
\end{figure}

\begin{figure}[t]
    \centering
    \includegraphics[width=0.88\linewidth]{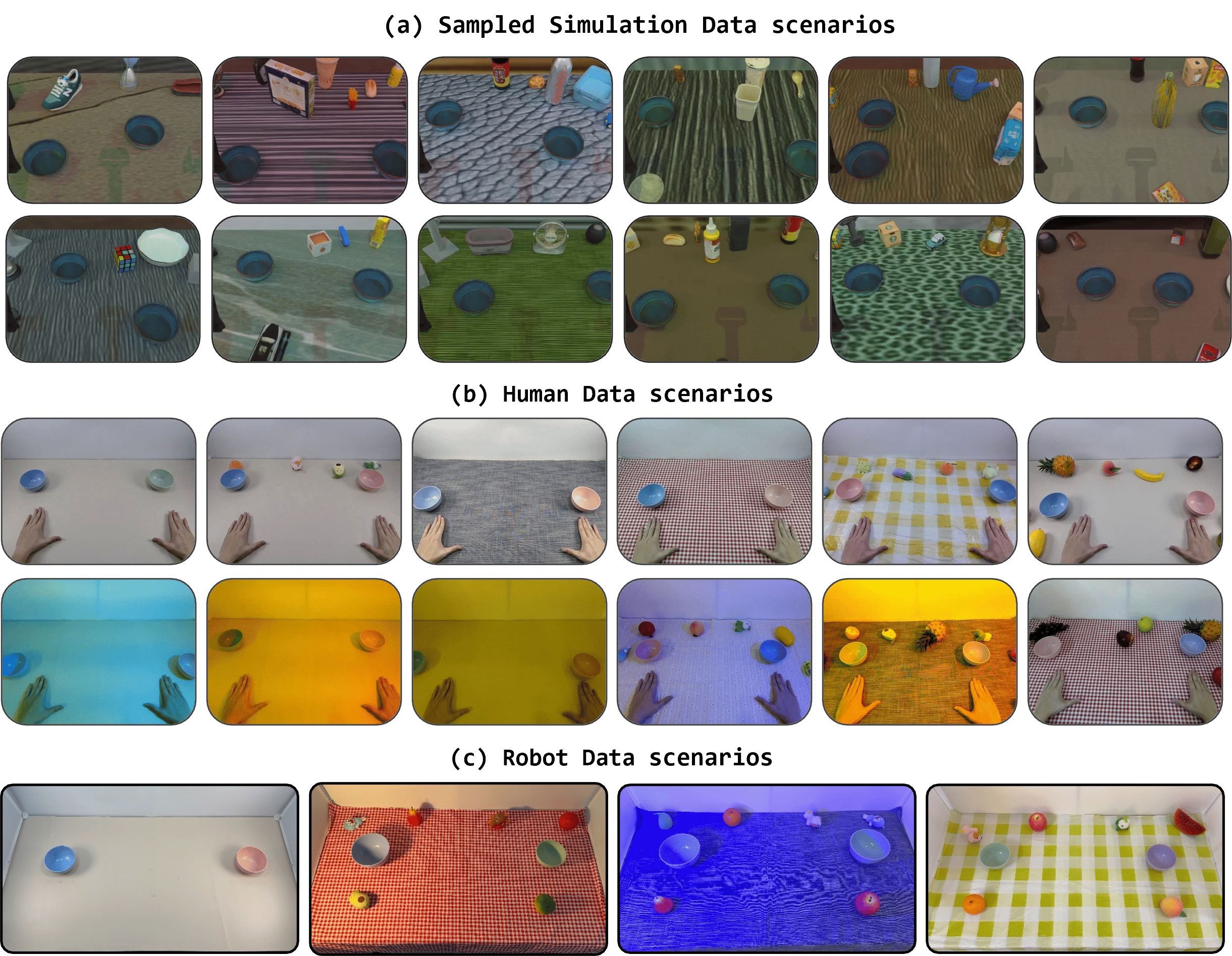}
    \caption{\textbf{Visualization of Data Collection Scenarios.} These diverse environments are constructed by systematically varying visual distractors ($\mathcal{F}_{dist}$), illumination ($\mathcal{F}_{light}$), and background ($\mathcal{F}_{bg}$). (a) Simulation environments leveraging extensive Domain Randomization to synthesize visual diversity. (b) Diverse human demonstration setups constructed by varying these environmental factors. (c) Real-robot setups ranging from simple, clean base scenarios to complex environments with significant clutter and lighting variations.}
    \label{fig:sim_human_data_scenarios}
\end{figure}

To evaluate robustness, we strictly define the environmental factors for each domain.
\begin{wrapfigure}[13]{r}{0.36\linewidth}
    \centering
    \vspace{-1em}
    \includegraphics[width=\linewidth]{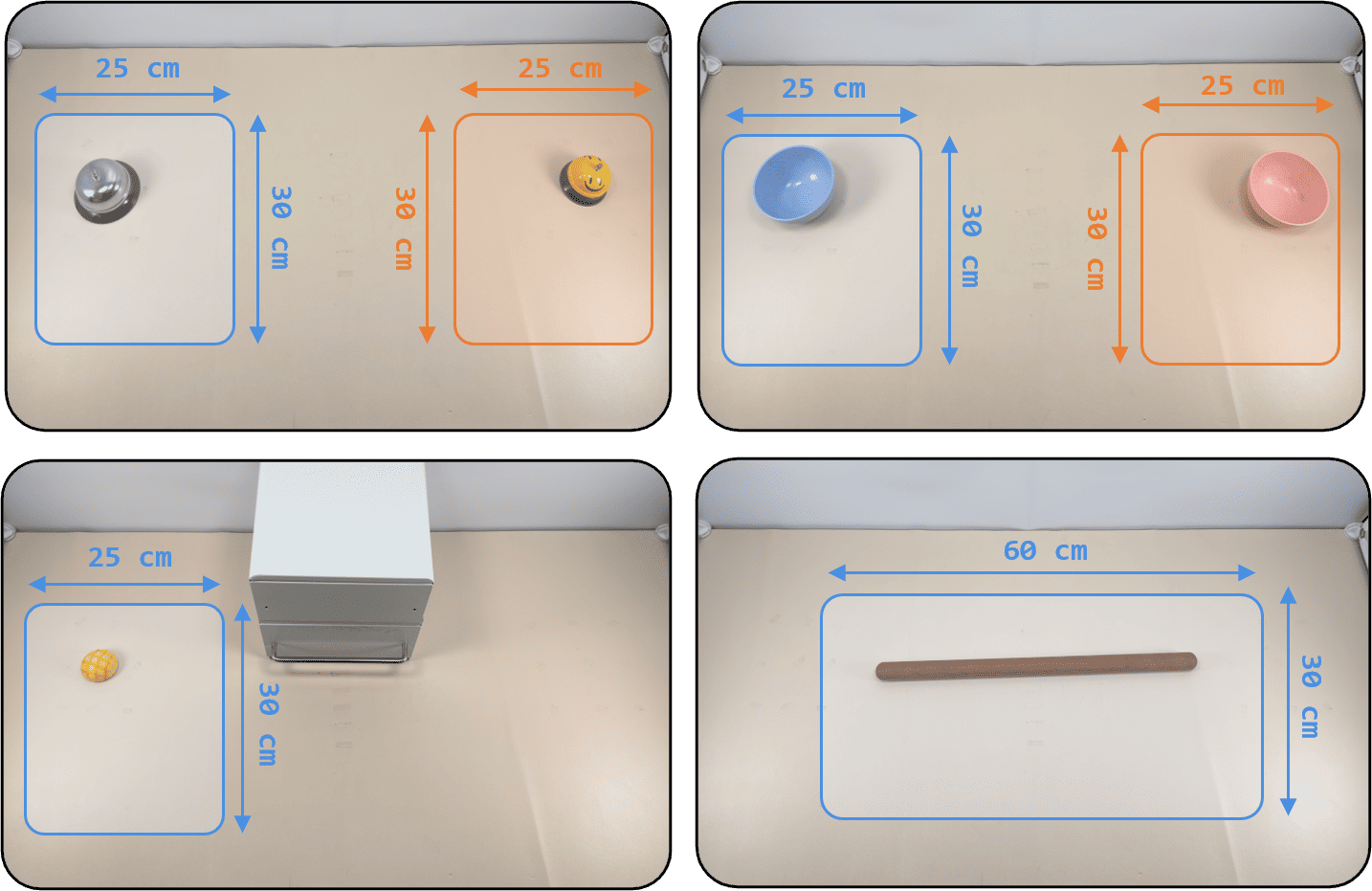}
    \caption{\textbf{Object Initialization Ranges.} Bounding boxes (in cm) show the spatial coverage for object placement ($\mathcal{F}_{init}$) across the four tasks.}
    \label{fig:init_position_setting}
    \vspace{-1em}
\end{wrapfigure}
The object instances ($\mathcal{F}_{obj}$) are visualized in Figure~\ref{fig:objects}, and valid spatial initialization ranges ($\mathcal{F}_{init}$) are shown in Figure~\ref{fig:init_position_setting}. We construct distinct scenarios by varying visual distractors ($\mathcal{F}_{dist}$), illumination ($\mathcal{F}_{light}$), and background ($\mathcal{F}_{bg}$). For $\mathcal{F}_{light}$, dynamic lighting is produced by a 40\,W photography flash lamp that strobes continuously throughout every rollout, independent of policy inference. \textbf{Simulation Scenarios:} we employ the built-in Domain Randomization engine of RoboTwin to randomize textures, lighting, and distractors (Figure~\ref{fig:sim_human_data_scenarios}(a)). \textbf{Human Scenarios:} to capture diverse real-world visual distributions, we curated 12 distinct scenarios by permuting environmental factors (Figure~\ref{fig:sim_human_data_scenarios}(b)), encompassing combinations of 4 tablecloth textures, 6 lighting conditions, and 16 distinct visual distractors to introduce realistic occlusion and clutter. \textbf{Real-Robot Scenarios:} to emulate a realistic data-scarce fine-tuning regime given the prohibitive cost of physical collection, we limited the design to 4 representative scenarios (1 base + 3 complex; Figure~\ref{fig:sim_human_data_scenarios}(c)) using combinations of 3 tablecloths, 3 lighting conditions, and 12 visual distractors.

\noindent\textbf{OOD factors and simulation randomization.} The OOD scenarios in Section~\ref{subsec:evaluation settings} hold out backgrounds, distractors, object instances, and lighting from the human and real-robot datasets. Simulation applies domain randomization, so we verify that these factors are also outside its coverage. The OOD objects are physical instances distinct from the synthetic assets (Figure~\ref{fig:objects}(a)). The OOD backgrounds share colors with the randomized textures but add irregular wrinkles that the renderer does not produce. The OOD distractors are fruits and plush toys, whereas the simulation distractors are rigid manipulation assets. The OOD lighting strobes within an episode, whereas RoboTwin only randomizes light color and intensity across episodes. The initial-pose factor is the exception: simulation samples positions over a broader range than the fixed grid used for human and real-robot collection, which is the point of Section~\ref{subsec: data contributions analysis}.

\subsection{Action Space and Data Processing}
\label{appendix:action_space}

We apply a standardized pipeline to align action representations across the three data sources, formulating all actions as relative trajectories~\citep{chi2024umi}.

\emph{Action representation.} For robot and simulation data, we record a 16-dimensional vector ($a_t^R \in \mathbb{R}^{16}$) comprising the 14-dimensional end-effector pose (position and quaternion) and binary gripper states. For human data, we record per hand a 7-DoF wrist pose (3D position and unit quaternion in the Meta Quest~3 world frame) plus the 3D offsets of five fingertips relative to the wrist position---22-dimensional per hand, concatenated to a 44-dimensional vector for both hands ($a_t^H \in \mathbb{R}^{44}$).\label{appendix:human_pose} Expressing fingertips as wrist-relative offsets keeps finger-shape information invariant to where the operator's wrist is located in the workspace. We do \emph{not} apply any explicit human-to-robot retargeting: the human and robot action spaces stay separate at the input/output layer via the modular encoders and decoders (Section~\ref{sec:model_arch}), and the shared backbone learns the embodiment mapping implicitly during pre-training.

\emph{Temporal alignment and pruning.} Human demonstrations are captured at 30\,Hz and treated as 10\,Hz sequences during training, effectively slowing the trajectory by 3$\times$ to compensate for the human-robot execution-speed gap. We also prune static segments at the beginning and end of each trajectory to remove operator-preparation idle periods.

\subsection{Data Collection Throughput}
\label{appendix:collection_throughput}

\begin{wrapfigure}[13]{r}{0.42\linewidth}
    \centering
    \vspace{-1em}
    \includegraphics[width=\linewidth]{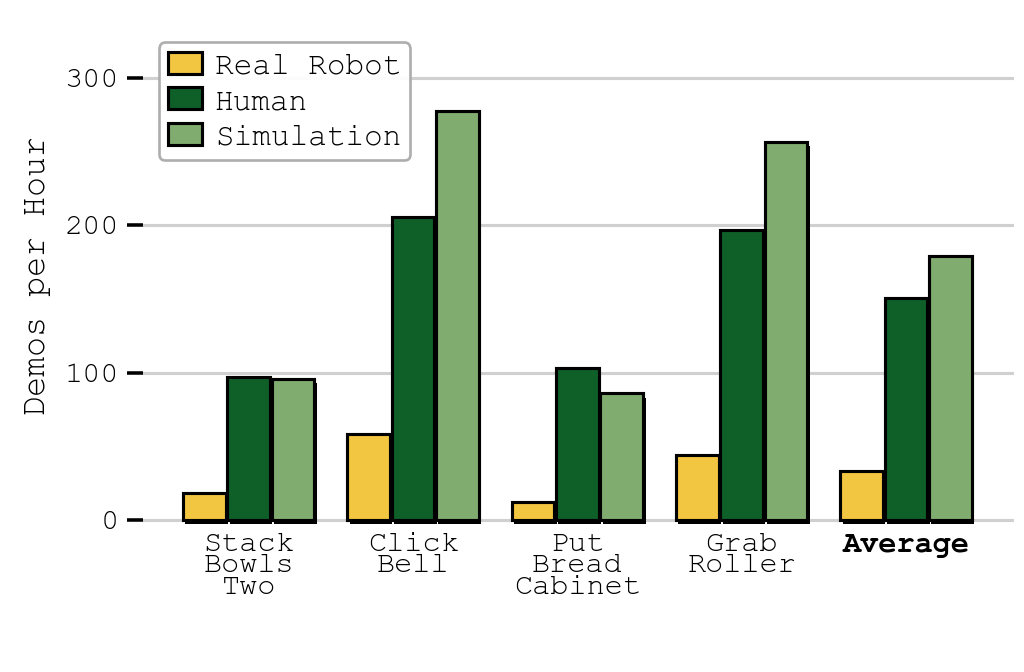}
    \caption{\textbf{Per-task collection throughput.} Demonstrations per hour across the four evaluation tasks; the ``Average'' column aggregates across tasks.}
    \label{fig:data_speed}
    \vspace{-1em}
\end{wrapfigure}
We compare demonstration collection speeds across the real-robot, human, and simulation pipelines; per-task results are reported in Figure~\ref{fig:data_speed}. Simulation throughput is measured with a single-threaded process on an NVIDIA RTX 4090 GPU. Human and real-robot measurements are averaged over operators, while the ``Average'' column in Figure~\ref{fig:data_speed} aggregates across tasks. On average, both simulation and human pipelines produce demonstrations roughly $\mathbf{5\times}$ faster than real-robot teleoperation and avoid real-robot operation during scalable data collection, making them useful sources for pre-training data. Generating the 500 simulation episodes used per task takes an estimated 3\,GPU-hours on a single RTX 4090 at the measured rate. Because simulation needs no operator and no robot, it runs in parallel with real-robot and human collection, and the budget study in Section~\ref{subsec:scaling} does not charge it against the collection budget of any method.

\section{Implementation Details and Hyperparameters}
\label{appendix:hyperparameters}

This appendix complements Section~\ref{sec:method} with full implementation details: visual encoder architecture (\S\ref{appendix:visual_encoder}), diffusion policy backbone (\S\ref{appendix:implementation_details}), and training hyperparameters and scope (\S\ref{appendix:training_hparams}).

We implemented our framework using the PyTorch library and managed training configurations via Hydra. Simulation data generation and policy training were run on a workstation equipped with 8 NVIDIA RTX 4090 GPUs.

\subsection{Visual Encoder Architecture}
\label{appendix:visual_encoder}

To process visual observations, we employed a ResNet-18 backbone initialized with ImageNet-1K pre-trained weights. We use independent visual encoders for each input camera view without weight sharing. The extracted feature maps are flattened, concatenated, and projected into the transformer's embedding dimension.

\subsection{Diffusion Policy Architecture}
\label{appendix:implementation_details}

Our policy architecture employs a shared DiT Backbone consistent with~\citep{dasari2024ditpi}. The diffusion process follows the Denoising Diffusion Probabilistic Models (DDPM), utilizing a squared cosine beta schedule with 100 diffusion steps during training, which is accelerated to 8 steps during inference. To enable unified training on heterogeneous data, we introduce Source-dependent Modules designed to capture the distinct characteristics of each domain. The implementation details of these modules are as follows:
\begin{itemize}
    \item \textbf{Domain-specific Vision Adaptors:} To process simulation and real-world inputs separately, we employ separate vision adaptors. Both adaptors are two-layer MLPs with GELU activations that project high-dimensional visual features into the latent space. Specifically, we utilize a Real-world Adaptor to process the embeddings from human ego-centric videos captured with a tripod-mounted RGB camera positioned to match the robot's head-camera viewpoint. In contrast, for the simulation domain, we instantiate independent Simulation Adaptors for each camera view to handle the multi-view synthetic data.
    \item \textbf{Modular Action Encoder:} We utilize separate projection layers to align heterogeneous kinematic spaces before they enter the shared backbone. Proprioceptive State Encoders project the robot state (16-dim) and human state (44-dim, same as the human pose) to the hidden dimension using a linear layer preceded by Dropout ($p=0.2$) to mitigate overfitting. Action Projectors encode the noisy action inputs during diffusion using a non-linear MLP ($\text{Linear} \to \text{GELU} \to \text{Linear}$) to effectively map the distinct action manifolds of human hands and robotic grippers into the shared latent space.
    \item \textbf{Modular Action Decoder:} The latent embeddings from the final backbone layer are projected into action space via separate heads. Specifically, distinct linear layers transform the shared features into 16-dimensional robot action noise and 44-dimensional human action noise, respectively.
\end{itemize}

\subsection{Training Hyperparameters and Scope}
\label{appendix:training_hparams}

\begin{table}[t]
\centering

\caption{Optimization hyperparameters.}
\label{tab:training_params}
\begin{tabular}{lcc}
\toprule
\textbf{Hyperparameter} & \textbf{Pre-train} & \textbf{Fine-tune} \\
\midrule
Training Steps & 200,000 & 60,000 \\
Optimizer & AdamW & AdamW \\
Learning Rate & $1 \times 10^{-4}$ & $5 \times 10^{-5}$ \\
Optimizer Betas & (0.95, 0.999) & (0.95, 0.999) \\
Weight Decay & $1 \times 10^{-6}$ & $1 \times 10^{-6}$ \\
LR Scheduler & Cosine Decay & Cosine Decay \\
Warmup Steps & 2000 & 2000 \\
\bottomrule
\end{tabular}
\vspace{-1em}
\end{table}
Table~\ref{tab:training_params} lists the optimization hyperparameters for Sim-and-Human pre-training and real-robot fine-tuning. Both stages are performed per task: pre-training draws mini-batches jointly from $\mathcal{D}_{sim}$ and $\mathcal{D}_{hum}$ at ratio $\alpha$ (Section~\ref{sec:training_recipe}); fine-tuning uses the corresponding 80-episode real-robot subset. Remaining low-level configuration details (batch size, DiT backbone width/depth, prediction and observation horizons, and the number of camera views per source) follow our reference DiT policy implementation~\citep{dasari2024ditpi}; we will release code and exact configurations to reproduce the experiments.\label{appendix:training_scope}

\section{Protocols for Decoupling and Scaling Experiments}
\label{appendix:exp_setup}

This appendix details the protocols underlying the decoupling analysis in Section~\ref{subsec: data contributions analysis} and the time-budget scaling study in Section~\ref{subsec:scaling}: human-data factor ablation (\S\ref{appendix:human_data_ablation_details}), position generalization (\S\ref{appendix:spatial_generalization_details}), and data-collection budget configurations (\S\ref{appendix:time_budget_details}).

\subsection{Human Data Factor Ablation}
\label{appendix:human_data_ablation_details}

\begin{figure}[t]
    \centering
    \includegraphics[width=0.84\linewidth]{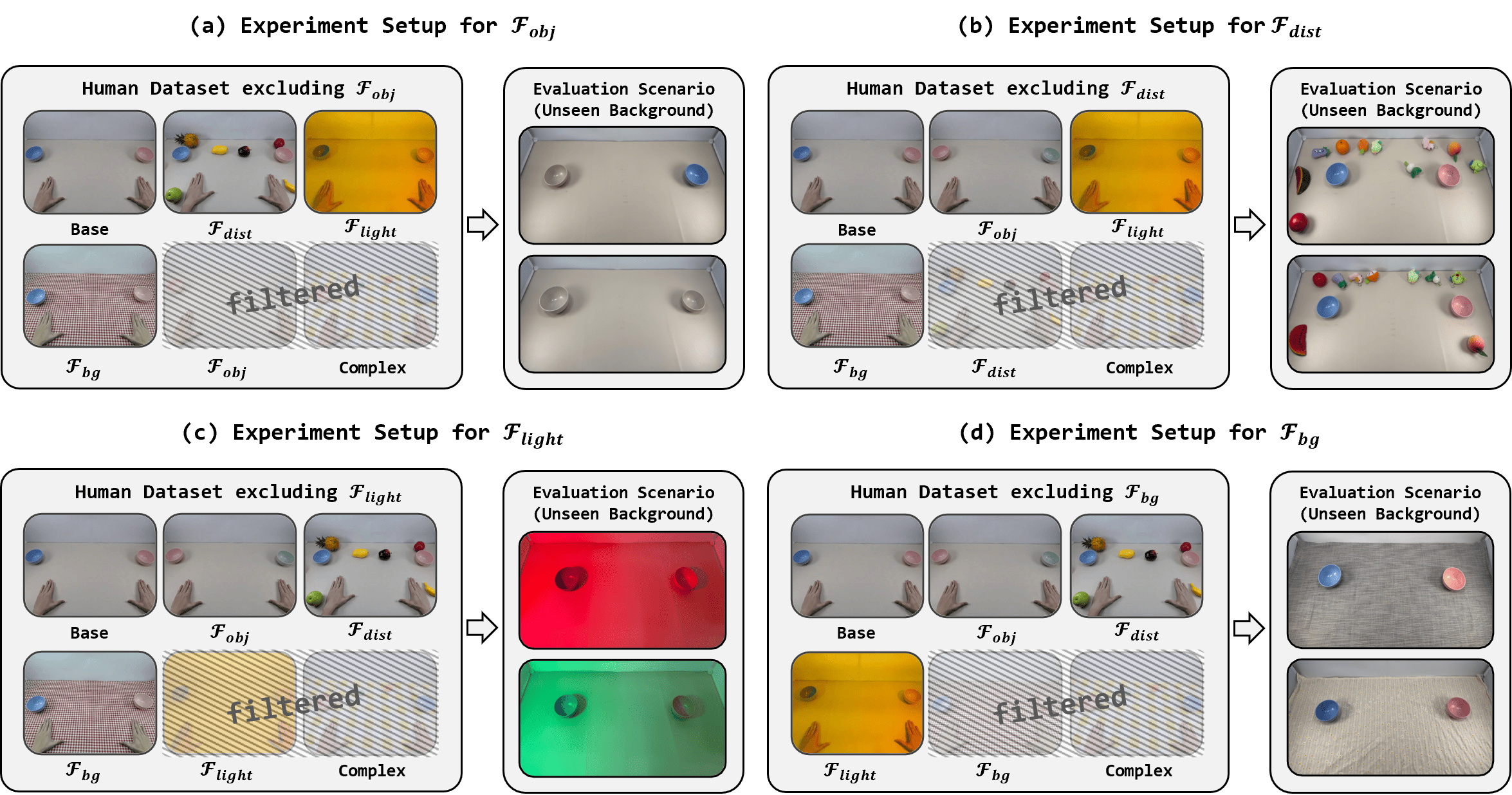}
    \caption{\textbf{Experiment Setup for Environmental Factor Ablation.} We systematically isolate the contribution of specific factors by curating training subsets that exclude data variations corresponding to a target factor (e.g., removing all tablecloth data for $\mathcal{F}_{bg}$). We then evaluate the model in targeted OOD scenarios that explicitly require robustness to the omitted factor, comparing performance against the full human dataset.}
    \label{fig:factor_ablation_setting}
\end{figure}

We detail the experimental protocol used to quantify the specific contributions of human data to visual generalization, as discussed in Section~\ref{subsec: data contributions analysis}. To measure the effect of each data factor ($\mathcal{F}_{bg}$, $\mathcal{F}_{dist}$, $\mathcal{F}_{light}$, $\mathcal{F}_{obj}$), we use a ``Leave-One-Factor-Out'' ablation strategy. The core logic of this experiment is illustrated in Figure~\ref{fig:factor_ablation_setting}. For each target factor, we construct a human-data subset that keeps the base condition for that factor while retaining variation in the other factors.

For a target factor (e.g., Background $\mathcal{F}_{bg}$), we first remove all human demonstrations that exhibit variations in that factor (e.g., excluding all trajectories collected on diverse tablecloths). The resulting training subset contains only the ``base'' condition for that specific factor (e.g., the default table surface) while retaining diversity in all other factors. Subsequently, we evaluate the policy trained on this restricted subset in unseen OOD scenarios that specifically feature the omitted factor. For instance, in the $\mathcal{F}_{bg}$ ablation, the model is evaluated on surfaces with novel textures (tablecloths) that were absent from its training distribution. Under this controlled comparison, degradation relative to the full-data baseline indicates sensitivity to missing visual diversity for that environmental factor.

\needspace{7\baselineskip}
\subsection{Position Generalization Analysis}
\label{appendix:spatial_generalization_details}

\begin{wrapfigure}[13]{r}{0.40\linewidth}
    \centering
    \vspace{-1em}
    \includegraphics[width=\linewidth]{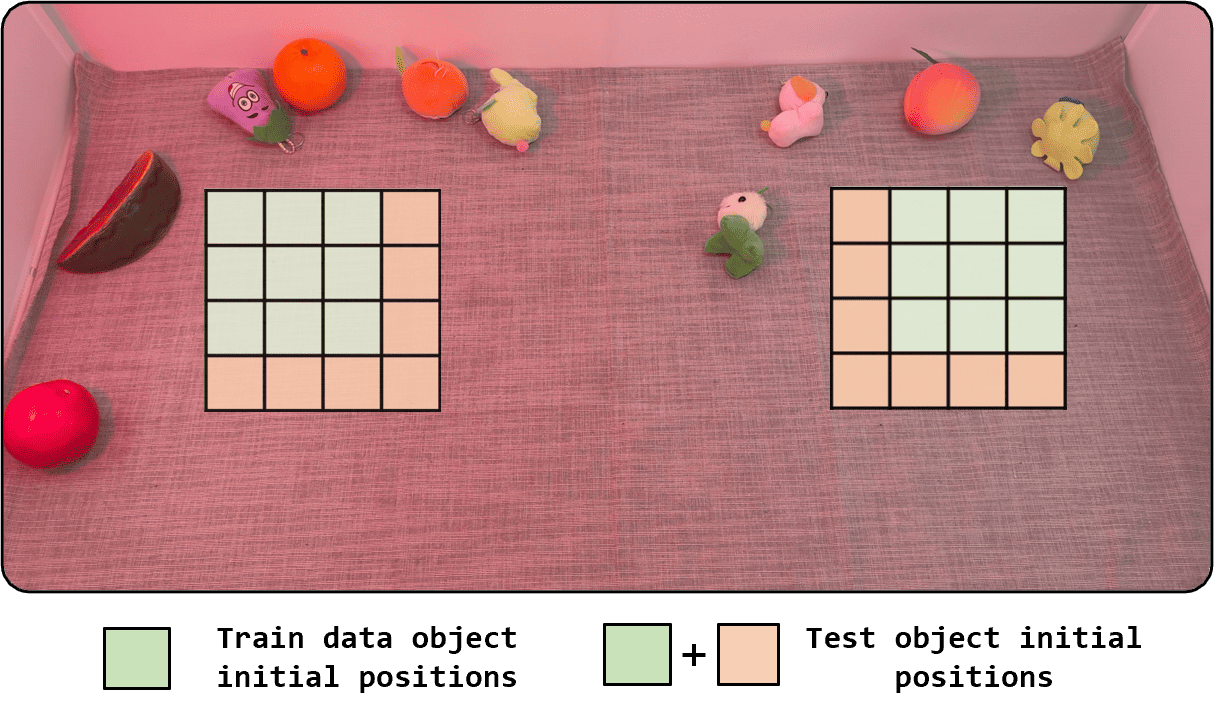}
    \caption{\textbf{Position Generalization Protocol.} Discretized $4{\times}4$ grids for Stack Bowls Two; training cells (Green) form the inner $3{\times}3$, evaluation extends to unseen outer cells (Orange).}
    \label{fig:position_generalization_setting}
    \vspace{-1em}
\end{wrapfigure}
We outline the experimental protocol used to evaluate the impact of simulation data on position generalization along the initial-pose factor ($\mathcal{F}_{init}$). We first collected 80 real-robot demonstrations for the Stack Bowls Two task within a pre-defined $3\times3$ spatial grid, utilizing the visual setups consistent with Figure~\ref{fig:sim_human_data_scenarios}(c). For evaluation, we introduce an OOD scene and expand the workspace to a larger $4\times4$ grid. Since the Stack Bowls Two task necessitates simultaneous manipulation in both left and right workspaces, we establish two symmetric $4\times4$ initialization grids to govern the object placement. As illustrated in Figure~\ref{fig:position_generalization_setting}, green zones denote the training coverage, while orange zones represent unseen positions. Evaluation is conducted across the full $4\times4$ grid, effectively probing the policy's performance on unseen outer coordinates (Orange). We conduct 10 independent evaluation trials for each of the 16 paired grid cells and report the average Progress Rate (PR).

\subsection{Time Budget Scaling Configurations}
\label{appendix:time_budget_details}

We provide the detailed data configurations for the Time Budget Scaling experiment discussed in Section~\ref{subsec:scaling}. Table~\ref{tab:time_budget_scaling_setting} lists the number of real-robot (R), simulation (S), and human (H) episodes assigned to each method under 2, 4, and 8 hour collection budgets. Simulation generation is automated and runs in parallel with human or real-robot collection, so it is not charged against the budget for any method. At 8\,h, SimHum therefore splits the budget evenly between real-robot teleoperation (80R) and human collection (400H) with 400S generated in parallel, whereas SimReal spends the full budget on real-robot teleoperation (160R) with the same 400S. The single-source baselines are evaluated at the 8\,h budget only.

\begin{table}[t]
    \centering
    \caption{\textbf{Data configurations for the Time Budget Scaling experiment.} The table lists the specific number of episodes collected by each method across different time budgets. The notations \textbf{R}, \textbf{S}, and \textbf{H} denote Real-world robot data, Simulation data, and Human data, respectively.}
    \label{tab:time_budget_scaling_setting}
    \begin{tabular}{l l l l}
        \toprule
        \textbf{Method} & \textbf{2h of data} & \textbf{4h of data} & \textbf{8h of data} \\
        \midrule
        Real only & 40R           & 80R           & 160R \\
        HumReal   & ---      & ---      & 80R+400H \\
        SimReal   & ---      & ---      & 160R+400S \\
        SimHum    & 19R+100S+100H & 40R+200S+200H & 80R+400S+400H \\
        \bottomrule
    \end{tabular}
\vspace{-8pt}
\end{table}

\section{Additional Results}
\label{appendix:additional_results}

\subsection{Full Main Results: Per-Task SR and PR}
\label{appendix:full_main_results}

Table~\ref{tab:main_results_full} reports the full per-task numbers complementing Figure~\ref{fig:main_results}, adding Progress Rate (PR) alongside Success Rate (SR) under In-Distribution (ID) and Out-of-Distribution (OOD) settings. Values are mean $\pm$ standard error.

\begin{table}[t]
\centering
\caption{\textbf{Full per-task SR and PR results across methods.} Mean $\pm$ standard error.}
\label{tab:main_results_full}
\setlength{\tabcolsep}{3pt}
\resizebox{\linewidth}{!}{%
\begin{tabular}{lcccccccccc}
\toprule
\multirow{2}{*}{\textbf{Method}} &
\multicolumn{2}{c}{\textbf{Stack Bowls Two}} &
\multicolumn{2}{c}{\textbf{Click Bell}} &
\multicolumn{2}{c}{\textbf{Put Bread Cabinet}} &
\multicolumn{2}{c}{\textbf{Grab Roller}} &
\multicolumn{2}{c}{\textbf{Average}} \\
\cmidrule(lr){2-3} \cmidrule(lr){4-5} \cmidrule(lr){6-7} \cmidrule(lr){8-9} \cmidrule(lr){10-11}
 & {\textbf{SR} $\uparrow$} & {\textbf{PR} $\uparrow$} &
   {\textbf{SR} $\uparrow$} & {\textbf{PR} $\uparrow$} &
   {\textbf{SR} $\uparrow$} & {\textbf{PR} $\uparrow$} &
   {\textbf{SR} $\uparrow$} & {\textbf{PR} $\uparrow$} &
   {\textbf{SR} $\uparrow$} & {\textbf{PR} $\uparrow$} \\
\midrule
\multicolumn{11}{c}{\textbf{In-Distribution Setting}} \\
\midrule
Real only & \acc{52.5}{7.9} & \acc{74.2}{7.0} & \acc{47.5}{7.9} & \acc{65.0}{8.0} & \acc{27.5}{7.1} & \acc{48.3}{8.0} & \acc{32.5}{7.4} & \acc{50.0}{8.0} & \acc{40.0}{3.9} & \acc{59.4}{3.9} \\
HumReal      & \acc{57.5}{7.8} & \acc{80.0}{6.0} & \acc{55.0}{7.9} & \acc{62.5}{8.0} & \acc{30.0}{7.2} & \acc{51.7}{8.0} & \acc{27.5}{7.1} & \acc{52.5}{8.0} & \acc{42.5}{3.9} & \acc{61.7}{3.8} \\
SimReal      & \acc{65.0}{7.5} & \acc{86.7}{5.0} & \acc{55.0}{7.9} & \acc{73.8}{7.0} & \acc{35.0}{7.5} & \acc{55.0}{8.0} & \acc{35.0}{7.5} & \acc{61.3}{8.0} & \acc{47.5}{3.9} & \acc{69.2}{3.6} \\
\rowcolor{lightcyan} \textbf{Ours} & \bfseries \acc{77.5}{6.6} & \bfseries \acc{90.0}{5.0} & \bfseries \acc{70.0}{7.2} & \bfseries \acc{83.8}{6.0} & \bfseries \acc{65.0}{7.5} & \bfseries \acc{78.3}{7.0} & \bfseries \acc{57.5}{7.8} & \bfseries \acc{76.3}{7.0} & \bfseries \acc{67.5}{3.7} & \bfseries \acc{82.1}{3.2} \\
\midrule
\multicolumn{11}{c}{\textbf{Out-of-Distribution Setting}} \\
\midrule
Real only & \acc{15.0}{8.0}  & \acc{48.3}{11.0} & \acc{15.0}{8.0}  & \acc{35.0}{11.0} & \acc{5.0}{4.9}   & \acc{23.3}{9.0} & \acc{0.0}{0.0}  & \acc{20.0}{9.0}  & \acc{8.8}{3.2}  & \acc{31.7}{5.1} \\
HumReal      & \acc{30.0}{10.2} & \acc{58.3}{11.0} & \acc{25.0}{9.7} & \acc{42.5}{11.0} & \acc{20.0}{8.9}  & \acc{45.0}{11.0} & \acc{15.0}{8.0} & \acc{27.5}{10.0} & \acc{22.5}{4.7} & \acc{43.3}{5.4} \\
SimReal      & \acc{40.0}{11.0} & \acc{60.0}{11.0} & \acc{40.0}{11.0} & \acc{52.5}{11.0} & \acc{20.0}{8.9}  & \acc{41.7}{11.0} & \acc{10.0}{6.7} & \acc{40.0}{11.0} & \acc{27.5}{5.0} & \acc{48.5}{5.5} \\
\rowcolor{lightcyan} \textbf{Ours} & \bfseries \acc{75.0}{9.7} & \bfseries \acc{83.3}{8.0} & \bfseries \acc{60.0}{11.0} & \bfseries \acc{80.0}{9.0} & \bfseries \acc{55.0}{11.1} & \bfseries \acc{68.3}{10.0} & \bfseries \acc{60.0}{11.0} & \bfseries \acc{77.5}{9.0} & \bfseries \acc{62.5}{5.4} & \bfseries \acc{77.3}{4.5} \\
\bottomrule
\end{tabular}%
}
\vspace{-8pt}
\end{table}

\subsection{Success Rates for Scaling Real-Robot Demonstrations}
\label{appendix:demo_scaling_sr}

Section~\ref{subsec:scaling} studies how OOD performance scales with the number of real-robot demonstrations, and Figure~\ref{fig:scaling}(b) plots Progress Rate. Figure~\ref{fig:demo_scaling_sr} reports Success Rate for the same study on Stack Bowls Two under OOD, with 20 trials per point. SimHum with 8 demos reaches the level of Real only with 160 demos in both metrics.

\noindent
\begin{minipage}[c]{0.40\linewidth}
\centering
\includegraphics[width=\linewidth]{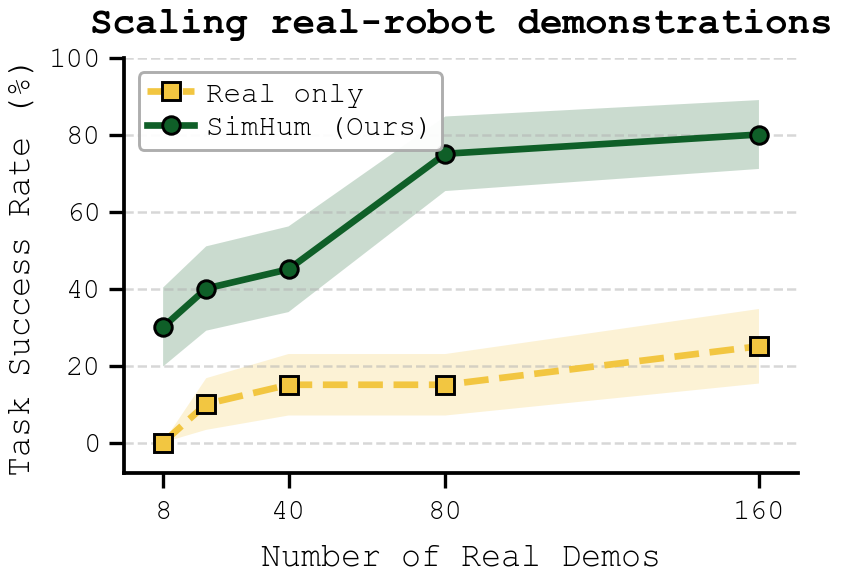}
\captionof{figure}{\textbf{Success Rate for scaling real-robot demonstrations.} Same setting as Figure~\ref{fig:scaling}(b).}
\label{fig:demo_scaling_sr}
\end{minipage}\hfill
\begin{minipage}[c]{0.56\linewidth}
\centering
\renewcommand{\arraystretch}{1.3}
\captionof{table}{\textbf{Additional comparisons on Stack Bowls Two (OOD, 20 trials).} All entries use the same 80 real-robot episodes for fine-tuning.}
\label{tab:additional_comparisons}
\vspace{2pt}
\small
\setlength{\tabcolsep}{4pt}
\begin{tabular}{lc}
\toprule
\textbf{Method} & \textbf{SR} $\uparrow$ \\
\midrule
$\pi_{0.5}$ fine-tuned~\citep{intelligence2025pi_} & \acc{55.0}{11.1} \\
EgoMimic-style co-training~\citep{kareer2025egomimic} & \acc{45.0}{11.1} \\
Single-stage sim+human+real co-training & \acc{55.0}{11.1} \\
\textbf{SimHum (Ours)} & \acc{75.0}{9.7} \\
\bottomrule
\end{tabular}
\end{minipage}

\subsection{Additional Comparisons}
\label{appendix:additional_comparisons}

Table~\ref{tab:additional_comparisons} compares SimHum against three alternatives on Stack Bowls Two under OOD. Each is fine-tuned on the same 80 real-robot episodes and evaluated with the same 20-trial protocol. \textbf{First,} fine-tuning $\pi_{0.5}$~\citep{intelligence2025pi_} on the same real-robot episodes matches SimHum in distribution (80.0\% vs.\ 77.5\% ID SR), indicating that the fine-tuning budget is sufficient for this backbone, but reaches 55.0\% under OOD against SimHum's 75.0\%. We attribute the OOD gap to the pre-training source. Simulation and, in particular, our self-collected human data lie much closer to the target domain than generic multi-embodiment pre-training, enabling extrapolation to the held-out factors that $\pi_{0.5}$'s generic priors do not cover. \textbf{Second,} an EgoMimic-style baseline~\citep{kareer2025egomimic} that co-trains on human and real-robot data without simulation reaches 45.0\%. In its rollouts the policy fails to grasp at some object positions, which we attribute to the position coverage that only simulation provides (Section~\ref{subsec: data contributions analysis}). \textbf{Finally,} mixing all three sources in a single stage at equal total compute, with real-robot data oversampled to a 1:1 ratio, reaches 55.0\% against 75.0\% for the staged recipe. We attribute this gap to interference between heterogeneous domain losses on the shared parameters, whereas the staged recipe ends with target-only adaptation on the deployment domain.

\subsection{Cross-Source Behavior Mismatch}
\label{appendix:behavior_mismatch}

Humans may perform a task differently from the robot, so human demonstrations can carry a behavior that the simulation and real-robot data do not. To probe whether such a mismatch hurts co-training, we replaced half of the human dataset with 250 newly collected demonstrations that grasp the bowl rim from four sides, a strategy absent from both the simulation and the real-robot data. OOD SR moves from 75.0\% to 65.0\%, within the standard error, and the fine-tuned policy still follows the grasp strategy of the real-robot demonstrations. This suggests that human data supplies visual priors rather than behaviors to imitate, and real-robot fine-tuning sets the policy's behavior.

\end{document}